\documentclass[journal]{IEEEtran}
\usepackage{amsmath,amssymb}
\usepackage{algorithmic}
\usepackage{graphicx}
\usepackage{textcomp}
\usepackage{epstopdf}
\usepackage{caption2}
\usepackage{algorithm}
\usepackage{subfigure}
\usepackage{dsfont}
\usepackage{amsfonts}
\usepackage{array}
\usepackage{booktabs}
\usepackage{indentfirst}
\usepackage{multirow}
\usepackage{color}
\usepackage{changepage}
\usepackage{lineno}
\usepackage[hang,flushmargin]{footmisc}
\usepackage{makecell,multirow,diagbox}
\usepackage{rotating}
\usepackage{hyperref}

\begin{document}
\title{Sparse Coding Driven Deep Decision Tree Ensembles for Nuclear Segmentation in Digital Pathology Images}

\author{Jie~Song,
        Liang~Xiao,~\IEEEmembership{Member,~IEEE,}
        Mohsen~Molaei,
        Zhichao~Lian,~\IEEEmembership{Member,~IEEE}
\thanks{{\kern 1pt} This work was supported in part by the National Natural Science Foundation of China under Grant 61871226, Grant 11431015, and Grant 61571230; in part by the National Major Research Plan of China under Grant 2016YFF0103604; in part by the Jiangsu Provincial Social Developing Project BE2018727; and in part by the Jiangsu Provincial Qinglan Project of Young and Middle-aged Academic Leaders. \emph{(Corresponding author: Liang Xiao.)}}
\thanks{{\kern 1pt} J. Song is with the College of Automation \& College of Artificial Intelligence, Nanjing University of Posts and Telecommunications, Nanjing 210023, China (e-mail: j.song0041@gmail.com; j.song@njupt.edu.cn).}
\thanks{{\kern 1pt} M. Molaei and Z. Lian are with the School of Computer Science and Engineering, Nanjing University of Science and Technology, Nanjing 210094, China (e-mail: mohsenmolaei@yahoo.com; newlzcts@gmail.com).}
\thanks{{\kern 1pt} L. Xiao is with the School of Computer Science and Engineering, Nanjing University of Science and Technology, Nanjing 210094, China, and also with the Key Laboratory of Intelligent Perception and Systems for High-Dimensional Information of Ministry of Education, Nanjing University of Science and Technology, Nanjing 210094, China (e-mail: xiaoliang@mail.njust.edu.cn).}}

\markboth{Journal of \LaTeX\ Class Files}%
{Shell \MakeLowercase{\textit{et al.}}: Bare Demo of IEEEtran.cls for IEEE Journals}

\maketitle

\begin{abstract}
In this paper, we propose an easily trained yet powerful representation learning approach with performance highly competitive to deep neural networks in a digital pathology image segmentation task. The method, called sparse coding driven deep decision tree ensembles that we abbreviate as Sc${{\bf{D}}^2}$TE, provides a new perspective on representation learning. We explore the possibility of stacking several layers based on non-differentiable pairwise modules and generate a densely concatenated architecture holding the characteristics of feature map reuse and end-to-end dense learning. Under this architecture, fast convolutional sparse coding is used to extract multi-level features from the output of each layer. In this way, rich image appearance models together with more contextual information are integrated by learning a series of decision tree ensembles. The appearance and the high-level context features of all the previous layers are seamlessly combined by concatenating them to feed-forward as input, which in turn makes the outputs of subsequent layers more accurate and the whole model efficient to train. Compared with deep neural networks, our proposed Sc${{\bf{D}}^2}$TE does not require back-propagation computation and depends on less hyper-parameters. Sc${{\bf{D}}^2}$TE is able to achieve a fast end-to-end pixel-wise training in a layer-wise manner. We demonstrated the superiority of our segmentation technique by evaluating it on the multi-disease state and multi-organ dataset where consistently higher performances were obtained for comparison against several state-of-the-art deep learning methods such as convolutional neural networks (CNN), fully convolutional networks (FCN), etc.
\end{abstract}


\IEEEpeerreviewmaketitle

\section{Introduction}
\IEEEPARstart{T}{he} automated segmentation of granular objects in digital pathology images remains one of the most challenging problems in cytology and neuropathology [\hyperlink{b1}{1}]-[\hyperlink{b3}{3}]. In such problems, nuclei in diverse images spanning a range of organs and disease states are expected to be precisely identified and segmented. Nevertheless, under pathological conditions, nuclei enlarge, exhibit margination of chromatin, and contain intra-nuclear sparse chromatin, prominent nucleoli are therefore identifiable; on the other hand, pathology images often exhibit high nuclear density, overlapping nuclear agglomerates, and background clutter with many artifacts, which exacerbate the difficulties of accurate nuclear segmentation. Conventionally, this problem can be roughly solved using a machine learning technique. Constructing a commonly known machine learning system requires careful engineering and considerable domain expertise to design a feature extractor that transforms the raw data into a suitable internal representation from which the system could segment patterns in the input, and thus they are not guaranteed to produce similar satisfactory performance in different scenarios. A comprehensive review of these techniques can be found in [\hyperlink{b4}{4}]--[\hyperlink{b6}{6}]. Therefore, the motivation of this work is the development of a new fully trainable system to learn meaningful representations of digital pathology data in an end-to-end fashion.

Before the introduction of representation learning, popular traditional methods use morphology operation, active contours, clustering, watershed transform, and their variants to solve microscopic granular object segmentation problems [\hyperlink{b7}{7}]--[\hyperlink{b10}{10}]. In order to accurately segment objects of interest, these methods require data-specific pre- and post-processing to obtain desired performance and so they do not directly generalize to segmentation problems across diverse nuclear types.

Currently, deep representation learning technique has provided an effective way of solving nuclear appearance diversity for pathology image segmentation [\hyperlink{b11}{11}]--[\hyperlink{b17}{17}] due to its ability to learn hierarchical abstract features from raw images without relying on prior knowledge. The two major implementations of such technique are convolutional neural networks (CNNs) [\hyperlink{b12}{12}], [\hyperlink{b13}{13}] and fully convolutional networks (FCNs) [\hyperlink{b14}{14}]--[\hyperlink{b17}{17}]. CNNs focus increasingly on drawing representational power from improved deeper architectures, such as Inception network [\hyperlink{b18}{18}], ResNet [\hyperlink{b19}{19}], and DenseNet [\hyperlink{b20}{20}], but still learn to map a fixed-size input to a fixed-size output. A pretrained CNN can be used as an encoder to generate a series of feature maps and further extended to derive a rich set of deep models that can be adapted to digital pathology image analysis applications, such as CNN2 [\hyperlink{b12}{12}] and CNN3 [\hyperlink{b13}{13}]. Compared with CNNs, FCNs can accept arbitrary-sized inputs and produce dense output predictions [\hyperlink{b21}{21}]. The strategy to achieve this goal is to add an effective decoder module to recover the spatial resolution from the output of encoder. For pathology image segmentation, the use of a variant of FCN, namely, U-Net [\hyperlink{b15}{15}] leads to significant improvement in accuracy by capturing and propagating contextual information to higher resolution layers with a \emph{skip connections} structure (see Figure 1 Third one). The success of faster R-CNN [\hyperlink{b22}{22}] is also extended to image segmentation problems [\hyperlink{b16}{16}], [\hyperlink{b17}{17}] by using a FCN for mask prediction and a ResNet-feature pyramid network (FPN) [\hyperlink{b23}{23}] backbone for feature extraction, where a top-down path with lateral connections is augmented to propagate semantically strong features. Although deep models are powerful, they have many deficiencies: 1) The networks are with too many hyper-parameters, and the performance depends seriously on careful parameter tuning. 2) Thus, they require extremely large amounts of labeled training data and computational complexity to reach state-of-the-art performance. 3) It is well known that deep models are black-box models whose decision processes are hard to understand, and the learning behaviors are very difficult for theoretical analysis [\hyperlink{b32}{32}].

Recently, several shallow learning-based lightweight models have been designed to tackle microscopy or pathology image segmentation problems (see Figure 1 Top and Second ones). The representative implementations of them include sparse coding [\hyperlink{b24}{24}], cascades of classifiers [\hyperlink{b25}{25}], [\hyperlink{b26}{26}], convolutional regression (CR) [\hyperlink{b27}{27}]--[\hyperlink{b30}{30}], and random forests [\hyperlink{b31}{31}], [\hyperlink{b32}{32}]. Sparse coding approach can greatly reduce the computational complexity without loss in accuracy by learning a set of separable 1D filters. Cascaded methods have much less hyper-parameters than deep models due to new classification formulations. CR techniques are much easier to train and in most cases, even when they are applied to different data across different domains, excellent performance can be achieved by almost same settings of hyper-parameters. A common strategy for some of these methods to achieve dense prediction is to extract a fixed-sized patch centered on each pixel and employ their learners to determine the label of the center pixel [\hyperlink{b25}{25}], [\hyperlink{b27}{27}], [\hyperlink{b31}{31}], [\hyperlink{b32}{32}]. However, such approaches only incorporate limited contextual information contained in the patch. The remaining models directly conduct the convolution operation on the whole input image and can recover the pixel-wise labels from their abstract feature representations [\hyperlink{b24}{24}], [\hyperlink{b26}{26}], [\hyperlink{b28}{28}]--[\hyperlink{b30}{30}]. However, they do not take into account the local dependencies between pixels, and thus there is still room for improvement in detailed feature reconstruction and compact representation. Thus, current pathology image segmentation problems face the conflicting goals of efficiency and accuracy.

An interesting recent attempt at automatic nuclear feature learning is the approaches of [\hyperlink{b33}{33}], [\hyperlink{b34}{34}], which rely on local sparse models designed to process overlapping patches independently. However, the independent sparse coding of each patch neglects the spatial correlation among them and leads to filters that are simply translated versions of each other. As a result, it generates highly redundant feature representation and results in a representation that is not optimal for the image as a whole. Convolutional sparse coding (CSC) [\hyperlink{b35}{35}] has been recently introduced as a global model to handle an entire image [\hyperlink{b26}{26}], [\hyperlink{b30}{30}]. The idea behind this strategy is to replace the linear combination of dictionary vectors by the sum of convolutions with dictionary filters. A disadvantage of this formulation is its computational expense. Although some authors resort to more efficient multiplications in the Fourier domain [\hyperlink{b36}{36}]--[\hyperlink{b38}{38}] or other efforts [\hyperlink{b26}{26}], [\hyperlink{b30}{30}], these attempts suffer from boundary condition limitations and ignore the local characterizations of the image. A different approach to Fourier domain implementations, i.e. slice-based dictionary learning [\hyperlink{b39}{39}], is currently developed for CSC, which uses a concatenation of banded circulant matrices and enables the solution of global problem in terms of only local computations in the original domain. Although the theoretical analysis of such fast CSC's success has been extended to several layers [\hyperlink{b40}{40}], leading to a new interpretation of CNNs, no work has been reported that they can achieve state-of-the-art performance in pathology image segmentation.

\begin{figure}[!t]
\renewcommand{\figurename}{\footnotesize{Fig.}}
\renewcommand{\captionlabeldelim}{.}
\centering
\includegraphics[width=8.5cm]{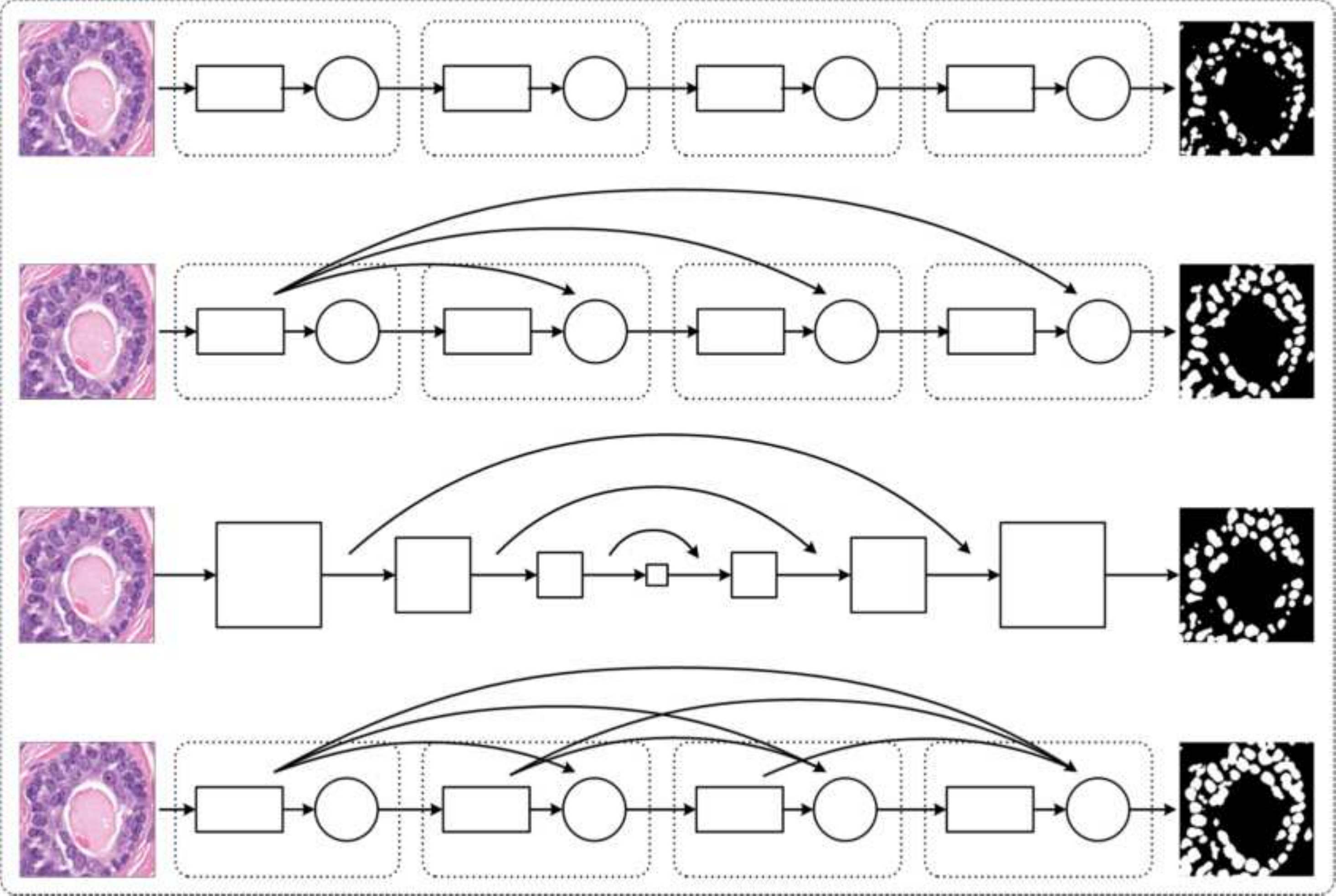}
\caption{\footnotesize{Different types of end-to-end learning structures for image segmentation. \textbf{Top row}: Basic CR structure without feature reuse (e.g. [28]), often results in coarse pixel masks since local appearance information is largely lost in the subsequent layers. \textbf{Second row}: The CR structure with low-level feature reuse (e.g. [30]) and \textbf{third row}: Fully convolutional network structure with low-level and high-level feature partly reuse (e.g. [15]). These two structures achieve impressive results on the digital pathology images, but the potentials of feature reuse are not deeply released. \textbf{Bottom row}: The proposed CR structure with all the low-level and high-level feature reuse. Concatenating the features learned by different layers increases the ability to collect more contextual information and also spatial information, and reduces the chance of assigning the pixels to the wrong positions.}}
\end{figure}

\begin{figure*}[!t]
\renewcommand{\figurename}{\footnotesize{Fig.}}
\renewcommand{\captionlabeldelim}{.}
\centering
\noindent\makebox[\textwidth][c] {
\includegraphics[width=17.8cm]{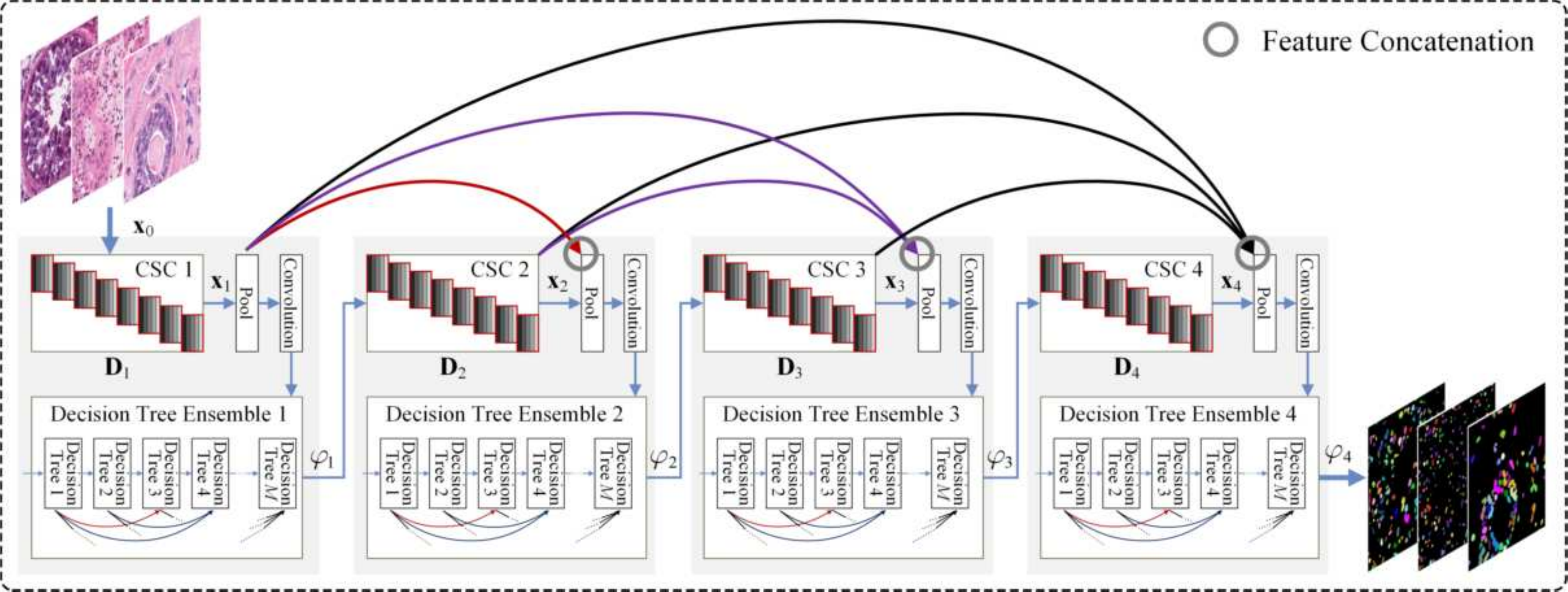}}
\caption{\footnotesize{Overview of Sc${{\rm{D}}^2}$TE with four layers. The feature maps extracted by the first layer from the raw image and the subsequent layers from score maps are fully reused. The Sc${{\rm{D}}^2}$TE combines the feature maps from all the previous fast CSC modules to form a new feature map pool as the input of subsequent decision tree ensemble module. After a fast CSC module, a 1$\times$1 convolution is introduced to compress the feature maps, followed by a decision tree ensemble module, aiming to learn $M$ simple and powerfully predictive decision trees. This module combines all the preceding optimal decision trees through summation before they are passed into a new iteration of tree learning and the pixel-wise regularized and complexity-constrained sum-of-squares loss is computed.}}
\vspace{-0.25cm}
\end{figure*}

In this work, we propose an easily trained yet powerful representation learning approach to address aforementioned challenges. Our method naturally balances the tradeoff between enhancing representation learning ability and controlling model complexity. We achieve these goals by reconstructing the convolutional regression structure and stacking multiple processing layers of CSC--decision tree ensemble pairwise modules to form a sparse coding driven deep decision tree ensembles (Sc${{\rm{D}}^2}$TE) structure (see Figure 1 Bottom row). The underlying idea of Sc${{\rm{D}}^2}$TE is based on the observation that if each layer is directly connected to every other layer in a feed-forward fashion then the model will be more accurate and easier to train. We choose fast CSC [\hyperlink{b39}{39}] as the feature detector in each layer, which benefits from having a local shift invariant structure and can efficiently find a global sparsest representation of the whole image or maps. We use an iterative additive process for regression until convergence, in which an ensemble mapping with random sampling, regularization, and pruning [\hyperlink{b41}{41}], [\hyperlink{b42}{42}] are applied progressively to all preceding feature maps to obtain a high order representation of the input, capture more contextual information, and improve predictive performance at each iteration. Sc${{\rm{D}}^2}$TE fully reuses the features of different levels through a sophisticated connectivity pattern, yielding condensed model that are easy to train, highly parameter-efficient, and without back-propagation. The optimal scoring of pixels being a nucleus is achieved by using only a small number of layers, thereby achieving the aforementioned conflicting goals. We show that the Sc${{\rm{D}}^2}$TE by itself, trained end-to-end, pixels-to-pixels, improve on the previous results in digital pathology nuclear segmentation. Experimental results on breast, prostate, kidney, stomach, and bladder pathology images demonstrate its superior performance to the state-of-the-art deep representation learning techniques and existing lightweight models.

\section{Sparse Coding Driven Deep Decision Tree Ensembles}
As shown in Figure 1, convolutional regression structures are built by alternatively stacking a feature extraction module and a prediction module, in which all the modules are subject to learning and compute non-linear input--output mappings. \emph{Skip connections} help the subsequent layers utilize multi-level features to improve predictive performance, by partly reusing feature maps. The goal of our model is to reinforce the concept of reuse by extending the multilayer stack of modules, while avoiding the feature explosion and reducing the overfitting on our segmentation task even with smaller training set sizes.

In this section, we introduce the proposed sparse coding driven deep decision tree ensembles (Sc${{\rm{D}}^2}$TE), as illustrated in Figure 2 comprehensively. We first present the overall model architecture. Next, fast CSC module for global representation learning and the decision tree ensemble module for latent structure prediction are explained in detail. At last, we discuss the problem of nuclear dense prediction using our Sc${{\rm{D}}^2}$TE.

\subsection{Overall Architecture of Sc${D}^2$TE}
We solve the segmentation problem by formulating it in terms of a densely concatenated convolutional regression problem. Consider an image ${{\bf{x}}_0} \in {\mathbb{R}}{^{n \times n}}$ that is passed through our Sc${{\rm{D}}^2}$TE, which comprises $L$ layers. The input of each decision tree ensemble module within a layer is the concatenation of all feature outputs of its previous layers after a compression with convolution operation, and the optimization of each decision tree within a decision tree ensemble module depends on all the previous learned decision trees. To preserve the feed-forward nature, each decision tree ensemble module then passes on its score output to the following fast CSC module of new layer.

Let ${{\bf{x}}_\ell} \in {\mathbb{R}}{^{n \times n}}$ be the feature output of the ${\ell}^{th}$ layer. In a basic convolutional regression model, ${\bf{x}}_\ell$ can be computed as follows: ${\bf{x}}_\ell={H_\ell}\left({\bf{x}}_{\ell-1}\right)$. ${H_\ell}\left( \cdot\right)$ is defined as a regression [\hyperlink{b27}{27}], [\hyperlink{b28}{28}] or iterative regression [\hyperlink{b29}{29}] followed by $c$ convolutions of size $q \times q$. Pushing this idea further, our Sc${{\rm{D}}^2}$TE introduces a sophisticated connectivity pattern that iteratively concatenates all feature outputs in a feed-forward fashion (see Figure 2). Thus, ${\bf{x}}_\ell$ is defined as
\begin{equation}
{\bf{x}}_\ell={H_\ell}\left( {\left[{{\bf{x}}_0},{{\bf{x}}_1},...,{{\bf{x}}_{\ell-1}} \right]} \right)
\end{equation}
where ${\left[{{\bf{x}}_0},{{\bf{x}}_1},...,{{\bf{x}}_{\ell-1}} \right]}$ denotes the concatenation operation of the feature maps ${{\bf{x}}_0},{{\bf{x}}_1},...,{{\bf{x}}_{\ell-1}}$. Meanwhile, ${H_\ell}\left( \cdot\right)$ is defined as a composite function of three consecutive operations: fast CSC, followed by a 1$\times$1 convolution, and a decision tree ensemble. Such connectivity pattern brings many advantages, such as strengthening feature propagation, encouraging feature reuse, and reducing the number of hyper-parameters.

From a different perspective, let $\varphi_l \in {\mathbb{R}}{^{n \times n}}$ be the score output of the ${\ell}^{th}$ layer. Our Sc${{\rm{D}}^2}$TE approach builds upon the observation that similar to the original image ${{\bf{x}}_0}$, the score map itself also admits a convolutional sparse representation. As such, it can be modeled as a superposition of filters, taken from a different global convolutional dictionary ${\bf{D}}_\ell$. Thus, we have
\begin{equation}
{{\bf{x}}_0}{\rm{ = }}{{\bf{D}}_1}{{\bf{x}}_1},\;{\varphi _1}{\rm{ = }}{{\bf{D}}_2}{{\bf{x}}_2},\;...,\;{\varphi _{L - 1}}{\rm{ = }}{{\bf{D}}_{L}}{{\bf{x}}_{L}}
\end{equation}
where ${\bf{D}}_\ell$ is a concatenation of $c$ banded circulant matrices, each of which has a band of width. Another elegant feature of our Sc${{\rm{D}}^2}$TE is that we can boost the predictive performance at each layer. Let ${{\bf{y}}_\ell} \in {\mathbb{R}}{^{n \times n}}$ be the target binary mask of the ${\ell}^{th}$ layer, and each point in ${{\bf{y}}_\ell}$ indicates the label of point in concatenated feature maps. Given training samples $\left( {\left[{{\bf{x}}_1},...,{{\bf{x}}_{\ell}} \right]},{{\bf{y}}_\ell} \right)$, the following decision tree ensemble method progressively provide an alternative to the activation function: $\varphi_\ell\left( \cdot\right)$ and fit current $\varphi_\ell\left( \cdot\right)$ to the training samples. More specifically at each iteration $m$, $\varphi_\ell^{m}\left( \cdot\right)$ first combines all the preceding learned decision trees $h_\ell^{1}\left( \cdot\right),h_\ell^{2}\left( \cdot\right),...,h_\ell^{m-1}\left( \cdot\right)$ through summation, and then grows a new decision tree $h_\ell^{m}\left( \cdot\right)$ by minimizing our loss function. This idea is illustrated in Figure 2 schematically and summarized below.
\begin{equation}
\varphi_\ell^{m}\left( {\left[{{\bf{x}}_1},...,{{\bf{x}}_{\ell}} \right]} \right) =  \varphi_\ell^{m-1}\left( {\left[{{\bf{x}}_1},...,{{\bf{x}}_{\ell}} \right]} \right) + h_\ell^{m}\left( {\left[{{\bf{x}}_1},...,{{\bf{x}}_{\ell}} \right]} \right)
\end{equation}
where $\varphi _\ell^{m - 1} = h_\ell^1 + h_\ell^2 +  \cdots  + h_\ell^{m - 1}$. The predictions from all decision trees are combined through a weighted majority vote to produce the final prediction. Note that we learn the $\varphi_\ell\left( \cdot\right)$ instead of the non-linear activation function, such that the back-propagation computation is not necessary for training.

\subsection{Multilayer Fast CSC for Global Representation Learning}
In the design of our CSC algorithm, we intend to achieve three goals: (1) Address the problem of training and using the CSC in the context of nuclear segmentation. (2) Extend such CSC approach to a much deeper configuration by assuming that the output of each layer itself also admits a convolutional sparse representation. (3) Improve the contextual information flow between layers through fully dense connectivity pattern.

We build on the fast CSC scheme proposed in [\hyperlink{b39}{39}] to learn a convolutional dictionary using the local-global decomposition, based on which our unsupervised compact feature learning module is constructed. Such CSC is capable of generating more complex filters capturing higher-older image statistics, compared to sparse coding that learns edge primitives.

\subsubsection{Fast CSC}
Assume each ${n \times n}$ pixel raw image can be represented by a vector ${{\bf{x}}_0} \in {\mathbb{R}}{^N}$. Our CSC module aims to decompose a ${{\bf{x}}_0}$ as ${{\bf{x}}_0} = {{\bf{D}}_1}{{\bf{x}}_1}$, where ${\bf{D}}_1$ is the concatenation of ${c_1}$ banded circulant matrices, each of which represents a convolution with one local $d_1$-dimensional filter. As such, a global convolutional dictionary ${\bf{D}}_1 \in {\mathbb{R}}{^{N \times N{c_1}}}$ is built from all shifted versions of a local dictionary ${{\bf{d}}_1} \in {\mathbb{R}}{^{{d_1} \times {c_1}}}$, containing the local filters as its columns, and the global sparse vector ${{\bf{x}}_1} \in {\mathbb{R}}{^{N{c_1}}}$ is obtained by simply interlacing all the feature maps $\left\{ {{\bf{x}}_{1,i}} \right\}_{i = 1}^{c_1}$. The intuition behind this description is given by Figure 3 Top. Using the above formulation, the task of CSC module amounts to solving the following problem
\begin{equation}
\mathop {\min }\limits_{{\bf{D}}_1,{{\bf{x}}_1}} \frac{1}{2}\left\| {{\bf{x}}_0 - {{\bf{D}}_1}{{\bf{x}}_1}} \right\|_2^2 + {\lambda _1} {\left\| {{\bf{x}}_1} \right\|_1}
\end{equation}
where the construction of global dictionary ${\bf{D}}_1$ is a balance between the reconstruction error and the ${\ell _1}$-norm penalty.

\begin{figure}[!t]
\renewcommand{\figurename}{\footnotesize{Fig.}}
\renewcommand{\captionlabeldelim}{.}
\centering
\includegraphics[width=8.6cm]{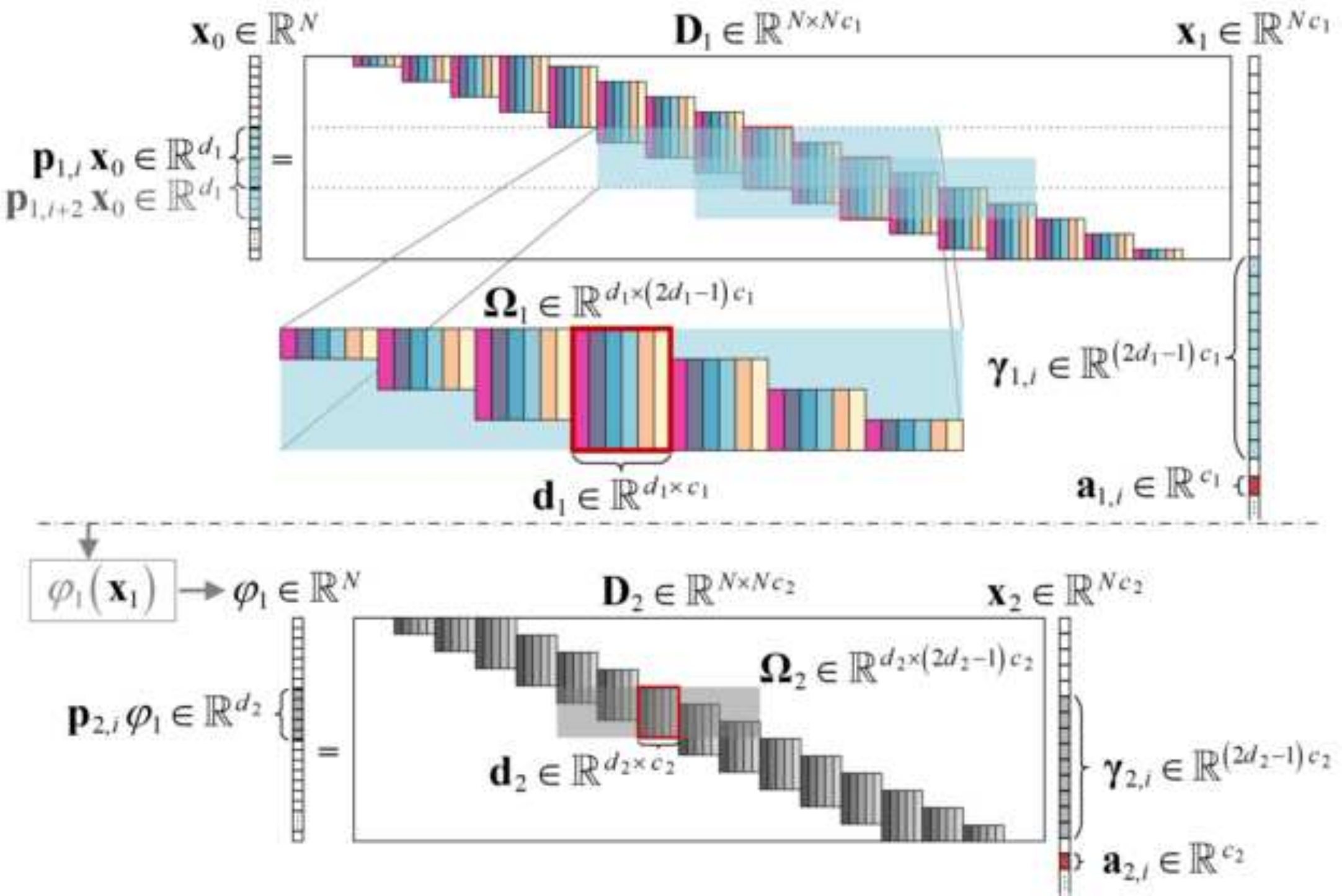}
\caption{\footnotesize{The fast CSC module (top), and its multilayer extension (bottom) by assuming that the score map also admits a convolutional sparse representation.}}
\vspace{-0.25cm}
\end{figure}

It is worth noting that the sparse vector ${{\bf{x}}_1}$ is built of $N$ distinct and independent local $c_1$-dimensional sparse codes ${\bf{a}}_{1,i} \in {\mathbb{R}}{^{c_1}}$, such that ${{\bf{x}}_0}$ can be expressed as ${{\bf{x}}_0} = \sum\nolimits_{i = 1}^N {{\bf{p}}_{1,i}^{\rm{T}}{{\bf{d}}_1}{\bf{a}}_{1,i}} $, where ${\bf{p}}_{1,i}^{\rm{T}} \in {\mathbb{R}}{^{N \times d_1}}$ is the operator that positions ${{\bf{d}}_1}{\bf{a}}_{1,i}$ in the $i^{th}$ location and pads the rest of entries with zeros. An immediate consequence of fast CSC module assumption is the fact that each patch ${{\bf{p}}_{1,i}}{{\bf{x}}_0} = {{\bf{p}}_{1,i}}{{\bf{D}}_1}{{\bf{x}}_1}$ taken from ${\bf{x}}_0$ can be expressed in terms of a shift-invariant local model ${{\bf{\Omega }}_1}{{\bf{\gamma }}_{1,i}}$, where ${{\bf{\Omega }}_1} \in {\mathbb{R}}{^{d_1 \times \left( {2{d_1} - 1} \right){c_1}}}$ is a \emph{stripe} dictionary containing ${{\bf{d}}_1}$ in its center, and ${{\bf{\gamma }}_{1,i}} \in {\mathbb{R}}{^{\left( {2{d_1} - 1} \right){c_1}}}$ is a \emph{stripe} vector, which can be seen as a group of $2{d_1}-1$ adjacent local $c_1$-dimensional sparse codes ${\bf{a}}_{1,j}$ from ${{\bf{x}}_1}$, centered at location ${\bf{a}}_{1,i}$. Using this local-global decomposition, the global optimization problem can be solved by alternatively optimizing over local sparse codes ${\bf{a}}_{1,i}$ and local dictionary ${{\bf{d}}_1}$:
\begin{equation}
\mathop {\min }\limits_{{{\bf{d}}_1},\left\{ {{{\bf{a}}_{1,i}}} \right\}_{i = 1}^N} \frac{1}{2}\left\| {{{\bf{x}}_0} - \sum\limits_{i = 1}^N {{\bf{p}}_{1,i}^{\rm{T}}{{\bf{d}}_1}{{\bf{a}}_{1,i}}} } \right\|_2^2 + {\lambda _1}\sum\limits_{i = 1}^N {{{\left\| {{{\bf{a}}_{1,i}}} \right\|}_1}}
\end{equation}
The minimization of (5) w.r.t. all the blocks $\left\{ {{{\bf{a}}_{1,i}}} \right\}_{i = 1}^N$ is separable, and can be solved sequentially for every ${{\bf{a}}_{1,i}}$ by using the LARS algorithm with local block coordinate descent [\hyperlink{b43}{43}].
Stochastic gradient descent is used for the minimization w.r.t. the local dictionary ${{\bf{d}}_1}$, which is trained requiring to constrain ${{\bf{d}}_1}$.
This two-stage optimization is then repeated until convergence. We kindly refer the reader to [\hyperlink{b43}{43}] for a more detailed description of mathematical derivation. Here we give the results of dictionary learning in this process in Figure 4(a).

\subsubsection{Multilayer Extension}
Interestingly, the success of this fast CSC approach has been extended in a hierarchical way [\hyperlink{b40}{40}], which provides an efficient theoretical framework for multi-level
feature extraction. The fast CSC assumes an inherent structure for raw images ${{\bf{x}}_0} \in {\mathbb{R}}{^N}$. Similarly, the score outputs $\varphi_\ell \in {\mathbb{R}}{^N}$ of each layer themselves could also be assumed to have such a structure after an ensemble mapping, e.g. ${\varphi _1}\left( {{{\bf{x}}_1}} \right)$. In what follows, we propose a multilayer fast CSC that relies on this rationale.

\begin{figure}[t]
\renewcommand{\figurename}{\footnotesize{Fig.}}
\renewcommand{\captionlabeldelim}{.}
\begin{center}
\includegraphics[width=8.6cm]{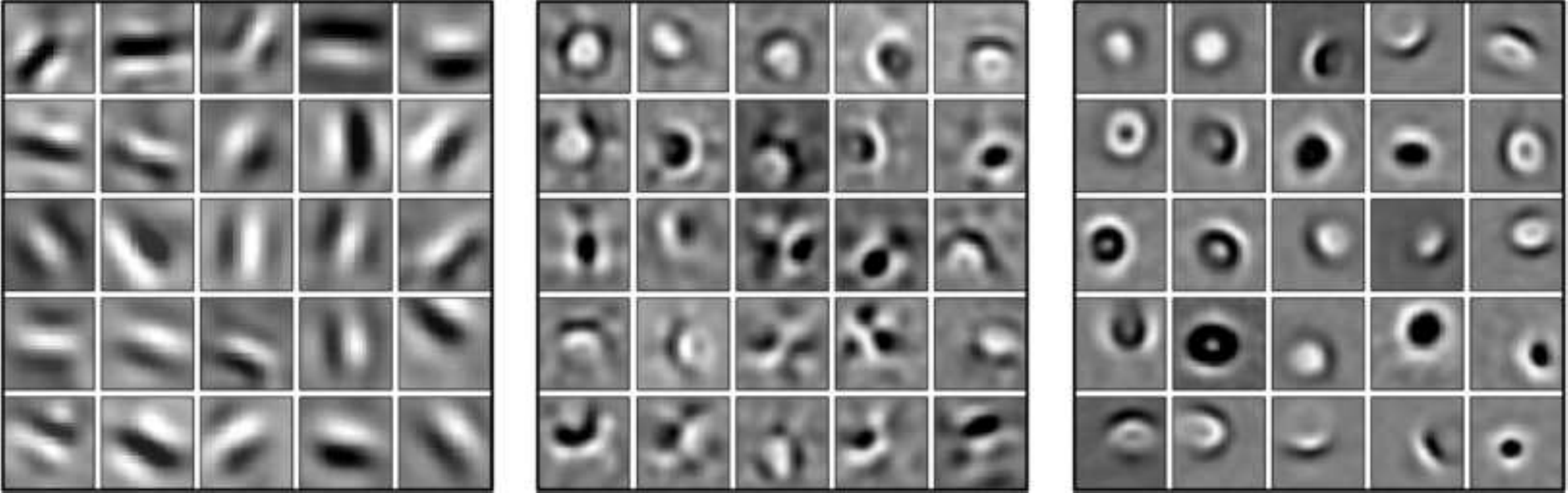}
\vspace{-0.3cm}
\small (a)\qquad\qquad\qquad\qquad(b)\qquad\qquad\qquad\qquad(c)
\end{center}
   \caption{\footnotesize{Multilayer fast CSC trained on the nuclear inputs. (a) $17 \times 17 $ local filters of dictionary ${\bf{D}}_1$ learned from raw images. (b) $29 \times 29 $ local filters of dictionary ${\bf{D}}_2$ effectively learned from the score output of first layer, and (c) of ${\bf{D}}_L$ effectively learned from the score output of ${\left(L-1\right)}^{th}$ layer.}}
\label{fig:long}
\label{fig:onecol}
\end{figure}

So far we have assumed ${{\bf{x}}_0}$ can be decomposed into a multiplication of a global convolutional dictionary ${\bf{D}}_1 \in {\mathbb{R}}{^{N \times N{c_1}}}$, composed of $c_1$ local $d_1$-dimensional filters ${{{\bf{f}}_{1,j}}}$, and a global sparse vector ${{\bf{x}}_1} \in {\mathbb{R}}{^{N{c_1}}}$. Herein, we extend this by proposing a similar factorization of vector ${\varphi _1}$, i.e., ${\varphi _1}{\rm{ = }}{{\bf{D}}_2}{{\bf{x}}_2}$, for a corresponding global convolutional dictionary ${\bf{D}}_2 \in {\mathbb{R}}{^{N \times N{c_2}}}$ with $c_2$ local $d_2$-dimensional filters ${{{\bf{f}}_{2,j}}}$ and a sparse representation ${{\bf{x}}_2} \in {\mathbb{R}}{^{N{c_2}}}$, as illustrated in Figure 3 Bottom. Under this multilayer construction, the vector ${\varphi _1}$ can be viewed both as a non-linearity with the sparse vector ${{\bf{x}}_1}$ as input or in itself an input that admits a sparse representation. Thus, the proposed multilayer fast CSC can be defined as
\begin{equation}
{\rm{find}}\; \left\{ {{{\bf{x}}_\ell}} \right\}_{\ell = 1}^L,\left\{ {{\bf{D}}_\ell} \right\}_{\ell = 1}^L\; {\rm{s}}{\rm{.t}}{\rm{.}}\; \forall \ell \;\;{\varphi _{\ell - 1}}{\rm{ = }}{{\bf{D}}_\ell}{{\bf{x}}_\ell},\;{\left\| {{\bf{x}}_\ell} \right\|_0} \le {{\rm{K}}_\ell}
\end{equation}
where we denote $\varphi_0$ to be the raw image ${{\bf{x}}_0}$ and ${{\rm{K}}_\ell}$ to be the cardinality constraint. Given a $\varphi_0$, this problem seeks for a set of representations $\left\{ {{{\bf{x}}_\ell}} \right\}_{\ell = 1}^L$ such that each one is locally sparse, and a set of dictionaries each of which is also sparse due to its unique structure. As mentioned above, the vector ${\varphi _{\ell-1}}$ is composed of $N$ non-overlapping local $c_\ell$-dimensional sparse codes ${\bf{a}}_{\ell,i} \in {\mathbb{R}}{^{c_\ell}}$, leading to the following definition of multilayer fast CSC:
\begin{flalign}
\begin{split}
& {\rm{find}}\; \left\{ {{{\bf{a}}_{\ell,i}}} \right\}_{i = 1}^N,\;{{\bf{d}}_\ell} = \left[ {{{\bf{f}}_{\ell,1}},{{\bf{f}}_{\ell,2}},...,{{\bf{f}}_{\ell,{c_\ell}}}} \right]\; {\rm{s}}{\rm{.t}}{\rm{.}}\; \forall \ell,i,j\\
& {\left\| {{\varphi _{\ell - 1}} - {\sum\limits_{i = 1}^N {{\bf{p}}_{\ell,i}^{\rm{T}}{{\bf{d}}_\ell}{{\bf{a}}_{\ell,i}}} } } \right\|_2} \le {\varepsilon_{\ell - 1}},\;{\left\| {{{\bf{a}}_{\ell,i}}} \right\|_0} \le {{\rm{k}}_\ell},\;{\left\| {{{\bf{f}}_{\ell,j}}} \right\|_2} = 1
\end{split}
\end{flalign}
where $\varepsilon_{\ell - 1}$ stands for the model mismatch. Intuitively, the fast CSC learns a superposition of atoms from $\varphi _0$ to ${\bf{x}}_1$. The multilayer version of this approach views $\varphi _0$ as a superposition of more complex entities, i.e. \emph{molecules}, from ${\varphi _1} = {\varphi _1}\left( {{{\bf{x}}_1}} \right)$ to ${\bf{x}}_2$, and \emph{cells} from ${\varphi _2} = {\varphi _2}\left( {{{\bf{x}}_2}} \right)$ to ${\bf{x}}_3$. Further layers continue to create more complex constructions and each adopts a different resolution of fundamental elements. We show an example of the resulting dictionaries in Figure 4.

\subsubsection{Feature Reuse}
Multilayer fast CSC exploits the property that digital pathology image are compositional hierarchies, in which higher-level context features are obtained by composing lower-level context ones. To further improve the contextual information flow between layers, we introduce a sophisticated connectivity pattern that densely concatenates all feature outputs $\left\{ {{{\bf{x}}_\ell}} \right\}_{\ell = 1}^L$ from previous layers, instead of just partly reusing the preceding feature outputs as in [\hyperlink{b15}{15}], [\hyperlink{b30}{30}]. This idea is summarized in (1). In addition of the feature pool generated by the dense concatenation, we also take into account the locations within a certain distance of each image pixel and add them into the pool for the purpose of exploiting additional image information. Each fast CSC feature map pool is followed by a 1$\times$1 convolution to compress the number and size of the feature maps and thereby reduces overfitting.

\subsection{Tree Ensemble Learning for Latent Structure Prediction}
The regression method we use to learn the features-to-score mapping in each layer is originated from Friedman \emph{et al.} [\hyperlink{b42}{42}]. To produce a series of simple and powerfully predictive decision trees, we make improvements in the regularized objective with the second order method [\hyperlink{b41}{41}], [\hyperlink{b44}{44}] and follow from the idea in dense connection to progressively update subsequent decision trees.

Let ${{\bf{b}}_{\ell,i}} = {\left[{{\bf{a}}_{1,i}},{{\bf{a}}_{2,i}},...,{{\bf{a}}_{\ell,i}} \right]}$ represents the concatenation of feature vectors ${{\bf{a}}_{1,i}},{{\bf{a}}_{2,i}},...,{{\bf{a}}_{\ell,i}}$, corresponding to the $i^{th}$ image location. For tree ensemble training, $t$ pairs of training samples $\left\{ {\left( {{{\bf{b}}_{\ell,i}},{y_{\ell,i}}} \right)} \right\}_{i = 1}^t$ are randomly sampled from the given data set. A decision tree ensemble module aims to learn $M$ simple and powerfully predictive decision trees $\left\{ {h_{\ell}^m\left(  \cdot  \right)} \right\}_{m = 1}^M :{{\mathbb{R}}^{{g_\ell}}} \mapsto {\mathbb{R}}, {g_\ell}>{c_\ell}$, and employ a weighted majority vote to predict the final score output of each layer:
\begin{equation}
{\varphi _\ell}\left({{\bf{b}}_{\ell}} \right) = \sum\limits_{m = 1}^M {\alpha _\ell^{m} h_\ell^{m} \left({{\bf{b}}_{\ell}} \right)}
\end{equation}
where $\alpha _\ell^{m} \in {\mathbb{R}}$ are contribution weights. In particular, the decision tree ensemble module tries to establish ${\varphi _\ell}\left(\cdot \right)$ that minimizes a loss ${\cal L} = {\cal L}\left( {{\bf{y}}_\ell},{\varphi_\ell^{m}\left( {{\bf{b}}_{\ell}} \right)}\right)$ by combining all the preceding learned decision trees through summation as follows
\vspace{-0.35cm}
\begin{equation}
\varphi_\ell^{m}\left( {{\bf{b}}_{\ell}} \right) = \varphi_\ell^{m-1} + \mathop {\arg \min }\limits_{h\left( \cdot \right)} {\cal L}\left( {{{\bf{y}}_\ell},\varphi _\ell^{m - 1} + \alpha h\left( {{{\bf{b}}_\ell}} \right)} \right)
\end{equation}
where $\varphi _\ell^{m - 1} \!=\! h_\ell^1 \!+\! h_\ell^2 \!+\! \cdots \!+\! h_\ell^{m - 1}$. Such dense connectivity pattern provides a useful extension of non-linear structures, making them more flexible while improving their predictive performance. More importantly, it yields a condensed model that is easy to train and highly parameter-efficient. Figure 5 shows the training process of such module.

\begin{figure*}[!t]
\renewcommand{\figurename}{\footnotesize{Fig.}}
\renewcommand{\captionlabeldelim}{.}
\centering
\noindent\makebox[\textwidth][c] {
\includegraphics[width=17.5cm]{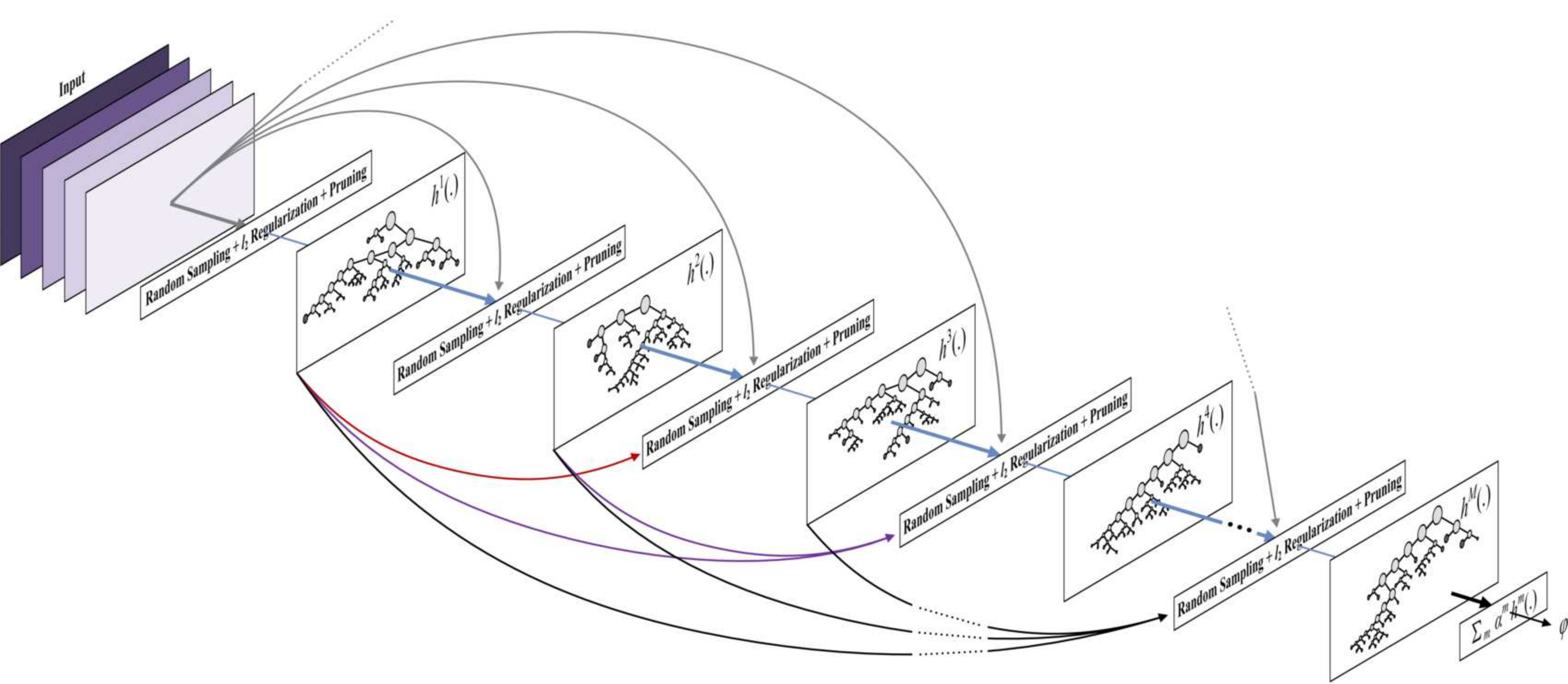}}
\caption{\footnotesize{Diagram of the training procedure for the decision tree ensemble module. At each ensemble iteration, we learn the decision tree using only a random subset of the whole training set. These trees are trained in sequence, and each new tree is built depending on all the previous learned trees, to minimize a loss $\cal L$ consisting of ${\ell _2}$-norm regularization and cost-complexity pruning. Once all the decision trees have been trained, their predictions are then combined through a weighted majority voting scheme. Note that, in each decision tree, the size of node is proportional to the percentage of features that reach this node.}}
\end{figure*}

We note that each decision tree ${h_{\ell}^m\left( \cdot \right)}$ can be defined by an independent structure and a set of leaf responses, such that
\begin{equation}
h_{\ell}^{m}\left( {{\bf{b}}_{\ell}} \right) = {{\bf{w}}_{s_\ell^{m}\left( {{\bf{b}}_{\ell}} \right)}}
\end{equation}
where ${s}_{\ell}^{m}\left( \cdot \right):{{\mathbb{R}}^{g_\ell}} \mapsto \left\{ {1,2,...,C_\ell^{m}} \right\}$ represents the structure of ${h_{\ell}^m\left( \cdot \right)}$ that maps an instance to the corresponding leaf index, where $C_\ell^{m} = |{h_{\ell}^m\left( \cdot \right)}|$ denotes the number of leaves in ${h_{\ell}^m\left(  \cdot \right)}$, and ${\bf{w}}_{s_\ell^{m}\left( \cdot \right)} = {\left( {{w_1},...,{w_{C_l^m}}} \right)^{\rm{T}}} \in {\mathbb{R}}{^{C_\ell^{m}}}$ is a vector of responses in leaves. For a given instance, we can thus use different tree structure to divide it into the leaves and calculate the prediction by summing up the responses in the corresponding leaves. To learn such $M$ decision trees in our decision tree ensemble module, we define the loss function ${\cal L}$ as follows
\begin{equation}
{\cal L} =
\left\| { {\bf{y}}_\ell - \left( {\varphi _\ell^{m - 1} + {{\bf{w}}_{s_\ell^{m}\left( {{\bf{b}}_{\ell}} \right)}}} \right) } \right\|_2^2 + \frac{1}{2}{\xi _\ell} \left\| { {{\bf{w}}_{s_\ell^{m}\left( {{\bf{b}}_{\ell}} \right)}} } \right\|_2^2 + {\zeta _\ell} C_\ell^{m}
\end{equation}
where the first term of (11) measures the closeness to the target, in which ${\bf{y}}_\ell = {\left( {y_{\ell,1}},{y_{\ell,2}},...,{y_{\ell,t}}\right)^{\rm{T}}}$, while the second and the third terms correspond to ${\ell _2}$-norm regularization and cost-complexity pruning [\hyperlink{b42}{42}], respectively, controlling the complexity of the module. Note that the coefficient ${\zeta _\ell}$ denotes the complexity cost by introducing additional leaf. Intuitively, minimizing the above loss function will help the algorithm to select simple and powerfully predictive decision trees. By defining ${\bf{B}}_{j} = \left\{ {i\left| {s_\ell^m\left( {{{\bf{b}}_{\ell,i}}} \right) = j} \right.} \right\}$ as the instance set of the $j^{th}$ leaf, we can use a second-order Taylor approximation to achieve fast optimization and rewrite (11) as
\begin{equation}
\begin{array}{l}
\vspace{0.15cm}
\tilde {\cal L}\left( {{y_{\ell,i}}, \varphi_\ell^{m}\left( {{\bf{b}}_{\ell,i}} \right)} \right) \simeq \sum\limits_{j = 1}^{C_\ell^{m}} {\left( { - 2{w_j}\left( {\sum\limits_{i \in {\bf{B}}_{j}} {\left( {{y_{\ell,i}} - \varphi _{\ell,i}^{{m - 1}}} \right)} } \right)} \right.} \\
{\kern 108pt} \left. { + w_j^2\left( {\sum\limits_{i \in {\bf{B}}_{j}} {1 + \frac{{{\xi _\ell}}}{2}} } \right)} \right) + {\zeta _\ell} C_\ell^{m}
\end{array}
\end{equation}
The local approximation approach of (12) to the loss function has faster convergence and gives a better approximation of tree structure and leaf response [\hyperlink{b41}{41}], [\hyperlink{b44}{44}]. For a fixed tree structure ${s}_{\ell}^{m}\left( \cdot \right)$, we therefore compute the optimal response ${\hat w_j}$ of the $j^{th}$ leaf via (12), as follows,
\begin{equation}
{\hat w_j} = \frac{{\sum\limits_{i \in {\bf{B}}_{j}} {\left( {{y_{\ell,i}} - \varphi _{\ell,i}^{{m - 1}}} \right)} }}{{\sum\limits_{i \in {\bf{B}}_{j}} {1 + \frac{{{\xi _\ell}}}{2}}}}
\end{equation}
where $\varphi _{\ell,i}^{m - 1} \!=\! h_{\ell,i}^1 + h_{\ell,i}^2 +  \cdots  + h_{\ell,i}^{m - 1}$. Once the ${\hat w_j}$ is obtained, we can then select the optimal decision tree structure $\hat {s}_{\ell}^{m}\left( \cdot \right)$ by calculating the optimal value of $\hat {\cal L}\left( {{y_{\ell,i}}, \varphi_\ell^{m}\left( {{\bf{b}}_{\ell,i}} \right)} \right)$:
\begin{equation}
\hat {\cal L}\left( {{y_{\ell,i}}, \varphi_\ell^{m}\left( {{\bf{b}}_{\ell,i}} \right)} \right) \!=\! - \sum\limits_{j = 1}^{C_\ell^{m}} {\frac{{{{\left( {\sum\limits_{i \in {\bf{B}}_{j}} {\left( {{y_{\ell,i}} \!-\! \varphi _{\ell,i}^{{m - 1}}} \right)} } \right)}^2}}}{{\sum\limits_{i \in {\bf{B}}_{j}} {1 + \frac{{{\xi _\ell}}}{2}}}}} \!+\! {\zeta _\ell} C_\ell^{m}
\end{equation}
The strategy for determining $\hat {s}_{\ell}^{m}\left( \cdot \right)$ is to use the greedy algorithm of [\hyperlink{b41}{41}], which starts from a single leaf and iteratively adds branches to the current decision tree. Let ${\bf{B}}^1$ and ${\bf{B}}^2$ represent the instance sets of left and right children after a split, respectively, and ${\bf{B}}  = {\bf{B}}^1 \cup {\bf{B}}^2$. Then, the loss reduction after the split: ${{\cal L}^{-}}$ is given by
\begin{equation}
\begin{array}{l}
\vspace{0.15cm}
{{\cal L}^{-}} = \frac{{{{\left( {\sum\limits_{i \in {\bf{B}}^1} {\left( {{y_{\ell,i}} - \varphi _{\ell,i}^{{m - 1}}} \right)} } \right)}^2}}}{{\sum\limits_{i \in {\bf{B}}^1} 1  + \frac{\xi _\ell}{2}}} + \frac{{{{\left( {\sum\limits_{i \in {\bf{B}}^2} {\left( {{y_{\ell,i}} - \varphi _{\ell,i}^{{m - 1}}} \right)} } \right)}^2}}}{{\sum\limits_{i \in {\bf{B}}^2} 1  + \frac{\xi _\ell}{2}}} \\
{\kern 111pt} - \frac{{{{\left( {\sum\limits_{i \in {\bf{B}} } {\left( {{y_{\ell,i}} - \varphi _{\ell,i}^{{m - 1}}} \right)} } \right)}^2}}}{{\sum\limits_{i \in {\bf{B}} } 1  + \frac{\xi _\ell}{2}}} - {\zeta _\ell}
\end{array}
\end{equation}
which is used to evaluate the split candidates. In particular, the first and the second terms of (15) correspond to the scores of left and right children, respectively, and the third term denotes the score if we do not split. Thus, the problem is converted into one of how to find the optimal splits. A feasible solution to this problem is to enumerate all the possible splits for continuous instances, but it is computationally expensive. Here we improve upon it by first sorting the instances based on their values and then mapping the similar ones to the corresponding splitting children. Finally, we accumulate the first-order and second-order statistics of left and right children after the split, respectively, and evaluate the current split through the method of (15). This algorithm terminates when ${{\cal L}^{-}} < 0$ or satisfying the predifined maximum depth and is repeated $M$ times.

At each ensemble iteration, it is worth noting that after obtaining the optimal decision tree $\hat {h}_{\ell}^{m}\left( \cdot \right)$, the weight $\alpha _l^{m}$ can be either calculated by a line search [\hyperlink{b30}{30}] or assigned to an average value of $\frac{1}{M}$. Then, the current module is updated using $\varphi _\ell^{m} = \varphi _\ell^{m-1} + h_{\ell}^{m}$, and passes on this result to next iteration to train a new decision tree. The overall procedure of decision tree ensemble algorithm is summarized in Algorithm 1. During its testing procedure, each new instance is independently pushed through the trained decision trees by using structures $\hat {s}_{\ell}^{m}\left( \cdot \right)$. When arriving at the leaves, the prediction of this instance is calculated by summing up the optimal values of ${\hat {w}_{j}}$ from different decision trees (see Figure 6).
\renewcommand{\algorithmicrequire}{\textbf{Input:}}
\renewcommand{\algorithmicensure}{\textbf{Output:}}
\begin{algorithm}[t]
\caption{Algorithm for Decision Tree Ensemble}
\label{alg1}
\hspace*{0.02in} {\bf Input:}
$t$ pairs of training samples:$\left\{ {\left( {{{\bf{b}}_{\ell,i}},{y_{\ell,i}}} \right)} \right\}_{i = 1}^t$, number \hspace*{0.07in}of decision trees $M$, and instance set ${\bf{B}}$ of current leaf.
\hspace*{0.02in} {\bf Output:}
${\hat \varphi _\ell} \equiv {\hat \varphi _\ell^{M}}$. \\
\hspace*{0.02in} {\bf Initialization:}
$\hat \varphi _\ell^{0} = {\hat \theta} = \mathop {\arg \min }\limits_\theta \sum\nolimits_{i = 1}^t {\left\| {{y_{\ell ,i}} - \theta } \right\|_2^2} $. \\
\vspace{-0.4cm}
\begin{algorithmic}[1]
\FOR{\emph{m} = 1 to \emph{M}}
   \STATE ${\cal G}_{\ell,i}^{m} = - 2\left( {{y_{\ell,i}} - \varphi _{\ell,i}^{{m - 1}}} \right)$, $G_\ell^{m} \leftarrow  - \frac{1}{2}\sum\nolimits_{i \in {\bf{B}} } {{\cal G}_{\ell,i}^{m}}$;
   \STATE ${\cal H}_{\ell,i}^{m} = 2$, $H_\ell^{m} \leftarrow \frac{1}{2}\sum\nolimits_{i \in {\bf{B}} } {{\cal H}_{\ell,i}^{m}}$;
   \STATE select a split that maximize ${{\cal L}^{-}}$ of (15) to determine the optimal structure $\hat {s}_{\ell}^{m}\left( \cdot \right)$ and $\{ \hat {\bf{B}} _j\} {_{j=1}^{\hat C_\ell^m}}$: \\
   ${{\cal L}^{-}} = \frac{{{{\left( {G_\ell^{m,1}} \right)}^2}}}{{H_\ell^{m,1} + \frac{{\xi _\ell} }{2}}} + \frac{{{{\left( {G_\ell^{m,2}} \right)}^2}}}{{H_\ell^{m,2} + \frac{{\xi _\ell} }{2}}} - \frac{{{{\left( {G_\ell^{m}} \right)}^2}}}{{H_\ell^{m} + \frac{{\xi _\ell} }{2}}} - {\zeta _\ell}$;
   \STATE For a given $\hat {s}_{\ell}^{m}\left( \cdot \right)$, determine leaf responses ${\{{\hat w_{j}} \} _{j=1}^{\hat C_\ell^m}}$: \\
   ${\hat {w}_{j}} = \frac{{G_\ell^{m}}}{{H_\ell^{m} + \frac{{\xi _\ell} }{2}}}$;
   \STATE $\hat h_\ell^{m} = \sum\nolimits_{j = 1}^{\hat C_\ell^m} {{{\hat w}_j}I\left( {i \in {{{\bf{\hat B}}}_j}} \right)}$;
   \STATE update $\hat \varphi _\ell^{m} = \hat \varphi _\ell^{m - 1} + \hat h_\ell^{m} = \hat h_\ell^{1} + \hat h_\ell^{2} + \cdots + \hat h_\ell^{m}$
\ENDFOR
\RETURN
${\hat \varphi _\ell^{M}} = \sum\nolimits_{m = 1}^M {\alpha _\ell^{m} \hat h_\ell^{m}}$
\end{algorithmic}
\end{algorithm}

\begin{figure}[!t]
\renewcommand{\figurename}{\footnotesize{Fig.}}
\renewcommand{\captionlabeldelim}{.}
\centering
\includegraphics[width=8.6cm]{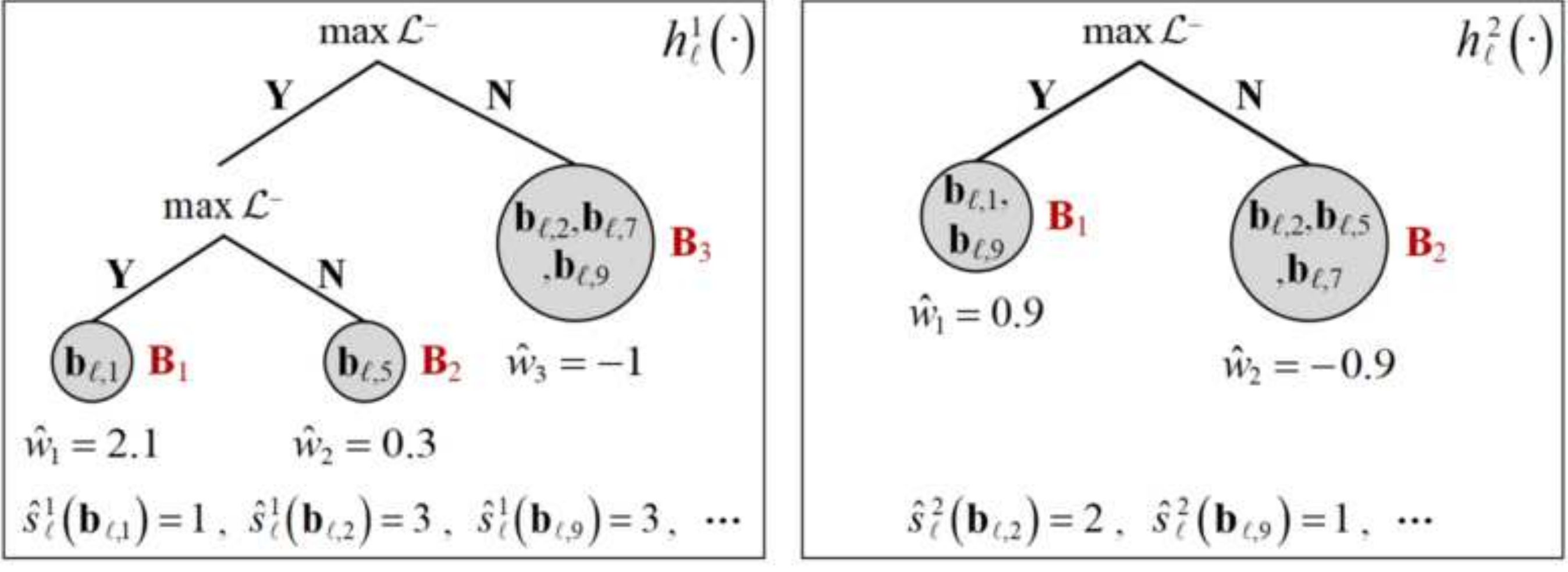}
\caption{\footnotesize{An example of the decision tree ensemble prediction. The contribution weight $\alpha _\ell^{m}$ is set to $\frac{1}{2}$. The final prediction for a given instance is the sum of predictions from each decision tree. For example, ${\hat \varphi _{\ell ,1}} = \frac{1}{2} \times 2.1 + \frac{1}{2} \times 0.9 = 1.5$, ${\hat \varphi _{\ell ,2}} = \frac{1}{2} \times \left( { - 1} \right) + \frac{1}{2} \times \left( { - 0.9} \right) =  - 0.95$, etc.}}
\vspace{-0.1cm}
\end{figure}

\subsection{Nuclear Dense Prediction Problem}
In our experiments, we used the dataset published in [\hyperlink{b13}{13}] for training and testing from different patients, organs, and disease states, containing more than 21,000 nuclei. To ensure richness of nuclear appearances, we covered breast, prostate, kidney, stomach, and bladder. Their whole slide images (WSIs) of size 1000$\times$1000 were enlarged to 200$\times$ on a 25" monitor such that each image pixel occupied 5$\times$5 screen pixels. The nuclei in the training set have been exhaustively annotated, generating a corresponding binary mask set of 12 high-resolution pathology images. The training set contains 11,460 nuclei that have been labeled across a wide spectrum of morphologies. Many nuclei exhibit margination and sparse properties of chromatin while the corresponding images comprises high-density agglomerates and background clutter with artifacts, which increase the complexity of the segmentation task.

\begin{figure}[!t]
\renewcommand{\figurename}{\footnotesize{Fig.}}
\renewcommand{\captionlabeldelim}{.}
\centering
\includegraphics[width=8.6cm]{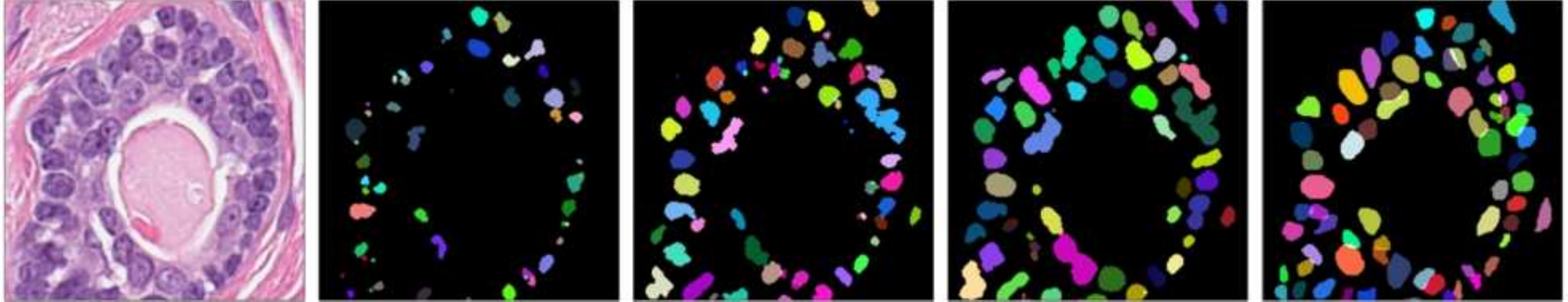}
\footnotesize{(a){\kern 40pt}(b){\kern 40pt}(c){\kern 40pt}(d){\kern 40pt}(e)}
\caption{\footnotesize{Improvement obtained by iterative regression. (a) Input image ${{\bf{x}}_0}$; (b) segmentation result ${\varphi _1}$ without (w/o) feature reuse; (c) segmentation result ${\varphi _2}$ with (w/) feature reuse; (d) segmentation result ${\varphi _L}$ with dense (w/d) feature reuse; (e) ground truth image ${\bf{y}}$. Both the iterative regression and dense feature reuse help to obtain a better segmentation accuracy. }}
\end{figure}

\begin{table}[!t]
\renewcommand\arraystretch{1.2}
\renewcommand{\captionfont}{\footnotesize}
\renewcommand{\captionlabeldelim}{}
\begin{center}
\caption {}
{\textsc{The Computed-Annotated Nucleus Matches for Our Method with Different Feature Map Reuses on the Testing Dataset}}
\vspace{0.2cm}
\setlength{\tabcolsep}{5pt}
\begin{tabular}{lccc}
\hline
Organ & w/o feature reuse & w/ feature reuse & w/d feature reuse \\
\hline
\emph{Same} & 1232 (47.5\%) & 1745 (67.3\%) & 2056 (79.3\%) \\
\emph{Different} & 1199 (62.8\%) & 1413 (74.1\%) & 1654 (86.7\%) \\
Overall & 2431 (54.0\%) & 3158 (70.2\%) & 3710 (82.4\%) \\
\hline
\end{tabular}
\label{tab1}
\end{center}
\vspace{-0.1cm}
\end{table}

We formulated this segmentation problem of 11,460 nuclei in the training set as a Sc${{\rm{D}}^2}$TE prediction problem. We built a pixel predictor with a layer-wise encoder-decoder structure (Section II) to progressively recover the latent nuclear regions. Each layer performed scoring by proceeding with a non-linear analysis on the random sample of nuclear features, which were generated using the ones from all the previous layers. This not only tackled the aforementioned challenges but took less time. The final nuclear segmentation was obtained by a $n$-ary mask for effective observation. Figure 7 shows the advantage of our approach on a sample nuclear image, and the results are shown in Table I. It is observed that the performance increases when feature maps are reused.

\begin{table}[!t]
\renewcommand\arraystretch{1.2}
\renewcommand{\captionfont}{\footnotesize}
\renewcommand{\captionlabeldelim}{}
\begin{center}
\caption {}
{\textsc{Our Dataset and Its Division for Training, Validation, and Same and Different Organs Testing}}
\vspace{0.2cm}
\setlength{\tabcolsep}{1.2pt}
\begin{tabular}{lccccccc}
\hline
\multirow{2}*{Dataset division} & \multicolumn{1}{c}{Nuclei} & \multicolumn{6}{c}{Images} \\
\cmidrule(r){3-8}
 & Total & Total & Breast & Prostate & Kidney & Stomach & Bladder \\
\hline
Train and validation & 11,460 & 12 & 4 & 4 & 4 & -- & -- \\
\emph{Same organ} test & 3,174 & 6 & 2 & 2 & 2 & -- & -- \\
\emph{Different organ} test & 3,393 & 4 & -- & -- & -- & 2 & 2 \\
\hline
\hline
Total & 18,027 & 22 & 6 & 6 & 6 & 2 & 2 \\
\hline
\end{tabular}
\label{tab2}
\end{center}
\vspace{-0.25cm}
\end{table}

\section{Experimental Results}
In this section, we first split the dataset into three sets to assess the performance of our approach, from which the best hyper-parameter settings of our Sc${{\rm{D}}^2}$TE are given. Then, we describe evaluation metrics and compare the segmentation obtained from our Sc${{\rm{D}}^2}$TE, U-Net [\hyperlink{b15}{15}], Mask R-CNN [\hyperlink{b16}{16}], CNN3 [\hyperlink{b13}{13}], multilayer CR (CR4) [\hyperlink{b30}{30}], single-layer CR (CR1) [\hyperlink{b28}{28}], and watershed (WS) [\hyperlink{b10}{10}]. We then evaluate the improvement brought by our method. Finally, we perform ablation analyses of Sc${{\rm{D}}^2}$TE. All the experiments are conducted on an assembly machine with one six-core Intel$^{\circledR}$ Core$^{\rm TM}$ (i7-8700K) (3.7GHz) and 32GB of memory.

\subsection{Training, Validation and Testing Set}
The Cancer Genome Atlas (TCGA) [\hyperlink{b45}{45}] is a publicly funded project that provides whole slide images (WSIs) of digitized tissue samples for nuclear segmentation. To maximize a diversity of nuclear appearances, we use the TCGA WSIs published in [\hyperlink{b13}{13}] for validation and comparison of the algorithms. We adopt a similar way and divide the whole dataset into three parts, as shown in Table II. The first part is used for training and validation. It corresponds to 12 patients and 3 organs including 4 breast, 4 prostate, and 4 kidney images with a total of 11,460 annotated nuclei. This amount is enough for us to train the deep models irrespective of whether they were implemented using the image-level classification or the pixel-wise prediction, even without data augmentation. The second part is used for the \emph{same organ} testing, which corresponds to the same organs as in the training set and contains 6 pathology images. We use this part to test the generalization of each method to three organs. The third part is added for \emph{different organ} testing. This part is more challenging because its images are taken from organs not represented in the training set, i.e. 2 stomach and 2 bladder images.

\subsection{Metrics}
As usually done to evaluate the segmentation methods [\hyperlink{b13}{13}], [\hyperlink{b30}{30}], we first introduce three local per pixel metrics:  Jaccard Index (JI), F1 score, and Average boundary distance (ABD). The F1 measure is defined as the harmonic mean between recall and precision at the pixel level and is equivalent to the dice coefficient. Let \emph{P}, \emph{R}, \emph{PB}, and \emph{RB} represent the output mask, ground truth mask, boundaries of \emph{P}, and boundaries of \emph{R}, respectively. The ABD measure is the average distance between \emph{PB} and \emph{RB}. These metrics are computed as follows
\vspace{0.1cm}
\begin{equation}
\begin{array}{l}
\vspace{0.15cm}
{\rm{JI}} =   \frac{{\left| {P \cap R } \right|}}{{\left| {P \cup R } \right|}}; \;
{\rm{F1}} =   \frac{2{\left| {P \cap R } \right|}}{{\left| {P } \right| + \left| {R } \right|}}; \\
\vspace{0.1cm}
{\rm{ABD}} {\rm{ = }} \frac{{{\rm{1}} }}{{\rm{2}}} \left( {\frac{{\sum\nolimits_{x \in PB} { d} \left( {x, RB} \right)}}{{\left| { PB } \right|}} + \frac{{\sum\nolimits_{u \in RB} { d} \left( {u, PB} \right) }}{{\left| { RB } \right|}}} \right) \\
\end{array}
\end{equation}

For the problem of nuclear segmentation, it is also interesting to have a global measure able to evaluate how well a region score can be used to recover the latent nuclear structure. To this end, we use the overlap (OV) measure that is similar to [\hyperlink{b29}{29}], which is defined as
\vspace{0.1cm}
\begin{equation}
{\rm{OV}} = \frac{{{\rm{TPR + TPM}}}}{{{\rm{TPR + TPM + FN + FP}}}}
\vspace{0.1cm}
\end{equation}
where TPR and TPM are the numbers of true positives in the reference nucleus and computed nucleus, respectively, and FN and FP are the numbers of false negatives and false positives.

\subsection{Models and Training Procedure}
For comparisons between ours and the state-of-the-arts, we studied three different deep models: U-Net, Mask R-CNN, and CNN3, and two lightweight models: CR4 and CR1. All the deep networks were trained using TensorFlow [\hyperlink{b46}{46}] on an NVIDIA GeForce GTX 1080 Ti$^{\circledR}$ graphics processing unit.

\subsubsection{Sc${{D}^2}$TE}
Given a set of model hyper-parameters, we train our Sc${{\rm{D}}^2}$TE with the images in the training set and evaluated the performance through the leave-one-out cross validation [\hyperlink{b47}{47}] on the validation set. This is repeated for all the sets of model hyper-parameters tested, and we finally choose the set of model hyper-parameters that optimizes the F1 score. The accuracy metrics we report are calculated on the testing sets containing the same and different organ images, none of which has been used neither for training nor for validation. Table III lists the results of the best hyper-parameter settings on the validation set. In our Matlab implementation and on a multicore machine, the average running time of our approach to predict the nuclear scores in an image is 0.034 hours.

\begin{table}[!t]
\renewcommand\arraystretch{1.2}
\renewcommand{\captionfont}{\footnotesize}
\renewcommand{\captionlabeldelim}{}
\begin{center}
\caption {}
{\textsc{The Best Hyper-Parameter Values of Our Model}}
\vspace{0.2cm}
\setlength{\tabcolsep}{5pt}
\begin{tabular}{lcccc}
\hline
Model & $L$ & ${d_\ell }\left( {{c_\ell }} \right)$ & $t$ & $M$ \\
\hline
Sc${{\rm{D}}^2}$TE & 4 & $17^2$,$29^2$,$29^2$,$29^2$ (32) & 50,000 & 30 \\
\hline
\end{tabular}
\label{tab3}
\end{center}
\vspace{-0.4cm}
\end{table}

\subsubsection{U-Net}
In our implementation of the U-Net architecture [\hyperlink{b15}{15}], we extracted image patches randomly, and did not perform data augmentation on them for computational reasons, but still achieved satisfactory performance. The U-Net was trained for 95 epochs. For its training, we set the batch size to 128 and we decay the learning every 10 epochs exponentially. We extracted 158,400 and 32,000 patches from 12 training and 3 validation images for training and validation, respectively. We tune the other hyper-parameters: the size and the number of filters in convolutional layers, strides, and upsampling factors such that the network can give a good performance on the held-out validation samples.

\subsubsection{Mask R-CNN}
The implementation of Mask R-CNN [\hyperlink{b16}{16}] is based on FPN and a ResNet. We set the backbone to ResNet 101 and the number of training anchors per image for the Region Proposal Network (RPN) part to 256. We set the number of RoIs per image to train with to 256. Furthermore, we use a batch size of 6 and decay the learning rate by 10 every 20 epochs. Mask R-CNN was trained for 80 epochs. We tuned the hyper-parameters: the Non-Maximum Suppression Threshold (NMST) and the Detection Minimum Confidence Rate (DMCR) for regions of interest. During training, it is better to have a high NMST as this means more proposals are generated and a high DMCR as this means more relevant examples are passed through the network.

\subsubsection{CNN3}
The CNN3 [\hyperlink{b13}{13}] was trained for 95 epochs. For its training, we set the batch size to 256 and decay the learning every 10 epochs exponentially. We tune the other hyper-parameters such as the size and the number of filters in convolutional layers, number of nodes in hidden layers, input-output size were selected, and dropout rate. We increased the dropout rate while going deeper in the network because a higher dropout rate in the initial layers results in information loss from the image but it acts as a good regularizer in the deeper layers to avoid overfitting.

\subsubsection{CR4\&CR1}
We trained CR4 [\hyperlink{b30}{30}] using 200,000 positive and negative samples, 100 weak regressors, and 4 model iterations. During training, a quarter of all the pixel locations were
randomly utilized to learn the model with 100 randomly selected local features at central pixel and 500 context features per channel. We also used the same features and the same parameters used for CR4 to train a CR1 model with one layer.

\subsubsection{WS}
We also experimented with the marker-based watershed transform [\hyperlink{b10}{10}], which implemented nuclear segmentation by using a Java-based Fiji plugin.

\begin{figure*}[!t]
\renewcommand{\figurename}{\footnotesize{Fig.}}
\renewcommand{\captionlabeldelim}{.}
\centering
\noindent\makebox[\textwidth][c] {
\includegraphics[width=17.5cm]{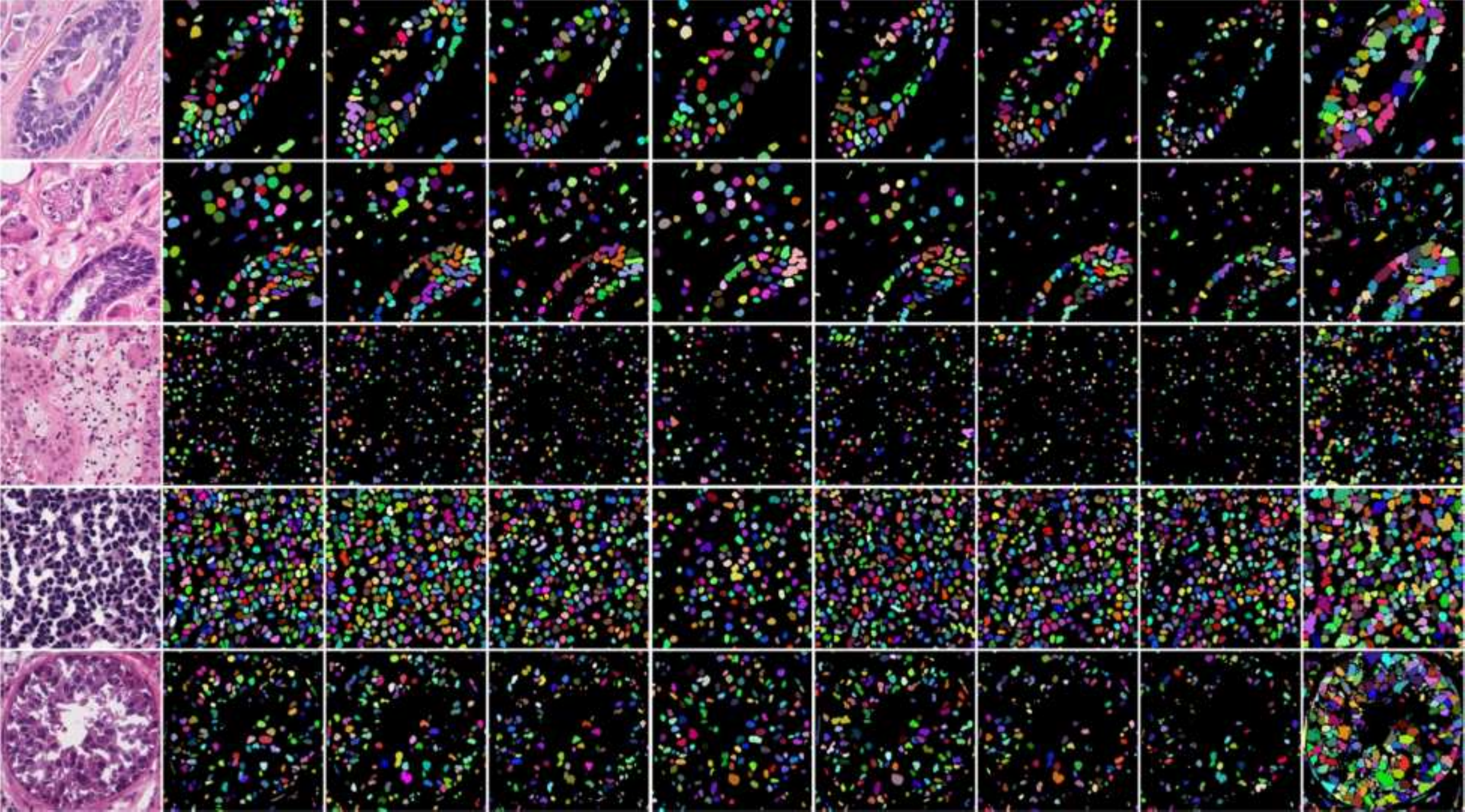}}
\footnotesize{(a){\kern 45pt}(b){\kern 45pt}(c){\kern 45pt}
(d){\kern 45pt}(e){\kern 50pt}(f){\kern 45pt}(g){\kern 45pt}(h){\kern 45pt}(i)}
\caption{\footnotesize{Comparative nuclear segmentation results on digital pathology images. (a) Raw images; (b) Ground truth images; (c) Segmentations generated by our method, (d) generated by U-Net [15], (e) generated by Mask R-CNN [16], (f) generated by CNN3 [13], (g) generated by CR4 [30], (h) generated by CR1 [28], and (i) generated by WS [10]. All the testing results are given in supplemental materials.}}
\end{figure*}

\begin{table*}[!t]
\renewcommand\arraystretch{1.19}
\renewcommand{\captionfont}{\footnotesize}
\renewcommand{\captionlabeldelim}{}
\begin{center}
\caption {}
{\textsc{Quantitative Analysis for Individual Images from Two Testing Sets in Terms of JI, F1, and ABD, in Which MRCNN Refers to Mask R-CNN [16]. The Top Results are Highlighted in {\textbf{\color[rgb]{0.4,0.4,1} Blue}} and Second-Best Results in {\textbf{\color[rgb]{0.6,0.8,0.2} Green}}}}
\vspace{0.2cm}
\setlength{\tabcolsep}{1.3pt}
\scriptsize
\begin{tabular}{lcccccccccccccccccccccccc}
\hline
\multirow{2}*{\bf Organ} & \multirow{2}*{\bf Image} & \multicolumn{7}{c}{\bf Jaccard Index} & \multicolumn{7}{c}{\bf F1 Score} & \multicolumn{7}{c}{\bf Average Boundary Distance} \\
\cmidrule(r){3-9} \cmidrule(r){10-16} \cmidrule(r){17-23}
      &  & Ours & U-Net & MRCNN & CNN3 & CR4 & CR1 & WS & Ours & U-Net & MRCNN & CNN3 & CR4 & CR1 & WS & Ours & U-Net & MRCNN & CNN3 & CR4 & CR1 & WS \\
\hline
\multirow{2}*{Breast} & {1} & {\bf \color[rgb]{0.4,0.4,1} 0.6267} & 0.5431 & 0.5720 & {\bf \color[rgb]{0.6,0.8,0.2} 0.6181} & 0.5353 & 0.3659 & 0.5407 & {\bf \color[rgb]{0.4,0.4,1} 0.7705} & 0.7039 & 0.7277 & {\bf \color[rgb]{0.6,0.8,0.2} 0.7640} & 0.6974 & 0.5358 & 0.7019 & {\bf \color[rgb]{0.4,0.4,1} 11.09} & 23.73 & 30.27 & {\bf \color[rgb]{0.6,0.8,0.2} 13.74} & 15.89 & 64.15 & 17.15 \\
       & {2} & 0.6468 & {\bf \color[rgb]{0.4,0.4,1} 0.6580} & {\bf \color[rgb]{0.6,0.8,0.2} 0.6512} & 0.5831 & 0.5805 & 0.4168 & 0.5821 & 0.7855 & {\bf \color[rgb]{0.4,0.4,1} 0.7937} & {\bf \color[rgb]{0.6,0.8,0.2} 0.7888} & 0.7367 & 0.7346 & 0.5883 & 0.7359 & {\bf \color[rgb]{0.6,0.8,0.2} 6.85} & {\bf \color[rgb]{0.4,0.4,1} 6.67} & 13.93 & 8.96 & 10.18 & 44.59 & 8.99 \\

\multirow{2}*{Prostate} & {1} & {\bf \color[rgb]{0.6,0.8,0.2} 0.5671} & 0.5001 & {\bf \color[rgb]{0.4,0.4,1} 0.6099} & 0.5290 & 0.3036 & 0.2601 & 0.4376 & {\bf \color[rgb]{0.6,0.8,0.2} 0.7238} & 0.6668 & {\bf \color[rgb]{0.4,0.4,1} 0.7577} & 0.6920 & 0.4658 & 0.4129 & 0.6088 & {\bf \color[rgb]{0.6,0.8,0.2} 17.74} & 19.10 & 19.22 & 22.24 & 69.02 & 65.85 & {\bf \color[rgb]{0.4,0.4,1} 16.78} \\
         & {2} & {\bf \color[rgb]{0.6,0.8,0.2} 0.6773} & 0.6607 & {\bf \color[rgb]{0.4,0.4,1} 0.6969} & 0.5553 & 0.5596 & 0.4286 & 0.6057 & {\bf \color[rgb]{0.6,0.8,0.2} 0.8076} & 0.7957 & {\bf \color[rgb]{0.4,0.4,1} 0.8214} & 0.7139 & 0.7176 & 0.6000 & 0.7544 & {\bf \color[rgb]{0.6,0.8,0.2} 6.76} & 8.19 & 13.36 & 29.17 & {\bf \color[rgb]{0.4,0.4,1} 6.57} & 17.92 & 13.25 \\

\multirow{2}*{Kidney} & {1} & {\bf \color[rgb]{0.4,0.4,1} 0.5520} & {\bf \color[rgb]{0.6,0.8,0.2} 0.5451} & 0.4197 & 0.5265 & 0.5252 & 0.3044 & 0.5270 & {\bf \color[rgb]{0.4,0.4,1} 0.7114} & {\bf \color[rgb]{0.6,0.8,0.2} 0.7056} & 0.5913 & 0.6898 & 0.6887 & 0.4667 & 0.6903 & {\bf \color[rgb]{0.4,0.4,1} 4.36} & 12.48 & 19.56 & 8.67 & {\bf \color[rgb]{0.6,0.8,0.2} 7.88} & 34.86 & 10.01 \\
       & {2} & {\bf \color[rgb]{0.4,0.4,1} 0.7146} & {\bf \color[rgb]{0.6,0.8,0.2} 0.6873} & 0.6349 & 0.5663 & 0.6517 & 0.5510 & 0.5688 & {\bf \color[rgb]{0.4,0.4,1} 0.8336} & {\bf \color[rgb]{0.6,0.8,0.2} 0.8147} & 0.7767 & 0.7230 & 0.7891 & 0.7105 & 0.7251 & {\bf \color[rgb]{0.6,0.8,0.2} 7.17} & 9.66 & 11.20 & 18.63 & {\bf \color[rgb]{0.4,0.4,1} 6.64} & 8.85 & 12.79 \\

\multirow{2}*{Stomach} & {1} & {\bf \color[rgb]{0.4,0.4,1} 0.6910} & 0.6545 & 0.4727 & 0.6247 & {\bf \color[rgb]{0.6,0.8,0.2} 0.6634} & 0.5254 & 0.6476 & {\bf \color[rgb]{0.4,0.4,1} 0.8173} & 0.7912 & 0.6420 & 0.7690 & {\bf \color[rgb]{0.6,0.8,0.2} 0.7977} & 0.6889 & 0.7861 & {\bf \color[rgb]{0.4,0.4,1} 1.09} & 3.97 & 19.74 & 4.27 & {\bf \color[rgb]{0.6,0.8,0.2} 1.10} & 2.73 & 10.04\\
        & {2} & {\bf \color[rgb]{0.4,0.4,1} 0.6838} & {\bf \color[rgb]{0.6,0.8,0.2} 0.6622} & 0.5164 & 0.5926 & 0.6405 & 0.5686 & 0.6560 & {\bf \color[rgb]{0.4,0.4,1} 0.8122} & {\bf \color[rgb]{0.6,0.8,0.2} 0.7968} & 0.6811 & 0.7442 & 0.7808 & 0.7250 & 0.7923 & {\bf \color[rgb]{0.4,0.4,1} 1.77} & {\bf \color[rgb]{0.6,0.8,0.2} 2.00} & 17.16 & 5.36 & 2.24 & 2.63 & 4.00 \\

\multirow{2}*{Bladder} & {1} & {\bf \color[rgb]{0.4,0.4,1} 0.6131} & 0.5504 & 0.5537 & {\bf \color[rgb]{0.6,0.8,0.2} 0.5835} & 0.5604 & 0.4855 & 0.3098 & {\bf \color[rgb]{0.4,0.4,1} 0.7602} & 0.7100 & 0.7128 & {\bf \color[rgb]{0.6,0.8,0.2} 0.7370} & 0.7183 & 0.6537 & 0.4731 & {\bf \color[rgb]{0.6,0.8,0.2} 16.15} & 31.09 & 26.78 & {\bf \color[rgb]{0.4,0.4,1} 13.79} & 35.83 & 41.74 & 48.27 \\
        & {2} & {\bf \color[rgb]{0.4,0.4,1} 0.7768} & 0.6730 & {\bf \color[rgb]{0.6,0.8,0.2} 0.7194} & 0.5808 & 0.6737 & 0.5980 & 0.4944 & {\bf \color[rgb]{0.4,0.4,1} 0.8744} & 0.8045 & {\bf \color[rgb]{0.6,0.8,0.2} 0.8368} & 0.7348 & 0.8050 & 0.7485 & 0.6617 & {\bf \color[rgb]{0.4,0.4,1} 3.30} & {\bf \color[rgb]{0.6,0.8,0.2} 6.13} & 8.36 & 10.46 & 7.76 & 10.77 & 13.50 \\
\hline
\hline
\bf \emph{Same} &  & {\bf \color[rgb]{0.4,0.4,1} 0.6308} & {\bf \color[rgb]{0.6,0.8,0.2} 0.5991} & 0.5974 & 0.5630 & 0.5260 & 0.3878 & 0.5436 & {\bf \color[rgb]{0.4,0.4,1} 0.7721} & {\bf \color[rgb]{0.6,0.8,0.2} 0.7467} & 0.7439 & 0.7199 & 0.6822 & 0.5524 & 0.7027 & {\bf \color[rgb]{0.4,0.4,1} 9.00} & 13.31 & 17.92 & 16.90 & 19.36 & 39.37 & {\bf \color[rgb]{0.6,0.8,0.2} 13.16} \\

\bf \emph{Different} &  & {\bf \color[rgb]{0.4,0.4,1} 0.6912} & {\bf \color[rgb]{0.6,0.8,0.2} 0.6350} & 0.5656 & 0.5954 & 0.6345 & 0.5444 & 0.5270 & {\bf \color[rgb]{0.4,0.4,1} 0.8160} & {\bf \color[rgb]{0.6,0.8,0.2} 0.7756} & 0.7182 & 0.7463 & 0.7755 & 0.7040 & 0.6783 & {\bf \color[rgb]{0.4,0.4,1} 5.64} & 10.86 & 18.01 & {\bf \color[rgb]{0.6,0.8,0.2} 8.47} & 11.73 & 14.47 & 18.95 \\

{\bf Overall} &  & {\bf \color[rgb]{0.4,0.4,1} 0.6549} & {\bf \color[rgb]{0.6,0.8,0.2} 0.6134} & 0.5847 & 0.5760 & 0.5694 & 0.4504 & 0.5370 & {\bf \color[rgb]{0.4,0.4,1} 0.7896} & {\bf \color[rgb]{0.6,0.8,0.2} 0.7583} & 0.7336 & 0.7304 & 0.7195 & 0.6130 & 0.6930 & {\bf \color[rgb]{0.4,0.4,1} 7.65} & {\bf \color[rgb]{0.6,0.8,0.2} 12.33} & 17.96 & 13.53 & 16.31 & 29.41 & 15.48 \\
\hline
\end{tabular}
\label{tab4}
\end{center}
\vspace{-0.2cm}
\end{table*}

\subsection{Qualitative Comparison of Methods}
{\color{black} Figure 8} shows a qualitative comparison between the results of aforementioned models on the multi-disease state and multi-organ pathology images. As seen, the traditional watershed method only works well for uniformly colored and isolated nuclei, such as kidney pathology nuclear segmentation. CR1 results in missed parts of the nuclear structures since only local information is used for prediction purposes. The CNN3 was limited by the image-level classification, which can also only extract local features. While the U-Net and CR4 can be better able to avoid the interference of complex background clutter by propagating contextual information to higher resolutions layers, it was still not sensitive to chromatin-sparse intra-nuclear details due to the small proportion of the feature reuse and the size of the network. As for the Mask R-CNN, the FPN and ResNet101 backbone made it to be the best network model to tackle segmentation problems involving sparse chromatin. However, due to the region proposal part of network, it did not generate enough positive proposals even for a high predefined NMST value. By contrast, clear advantages of our Sc${{\rm{D}}^2}$TE are that it was immune to the nuclear appearance diversity, can effectively deal with variations on nuclear size, shape, and direction, and detect the correct number of nuclei with high accuracy. Furthermore, for the case of sparse chromatin and heavy background clutter (such as might occur in prostate and bladder pathology images), the proposed Sc${{\rm{D}}^2}$TE also exhibits robustness.

\begin{figure*}[!t]
\renewcommand{\figurename}{\footnotesize{Fig.}}
\renewcommand{\captionlabeldelim}{.}
\centering
\noindent\makebox[\textwidth][c] {
\includegraphics[width=18cm]{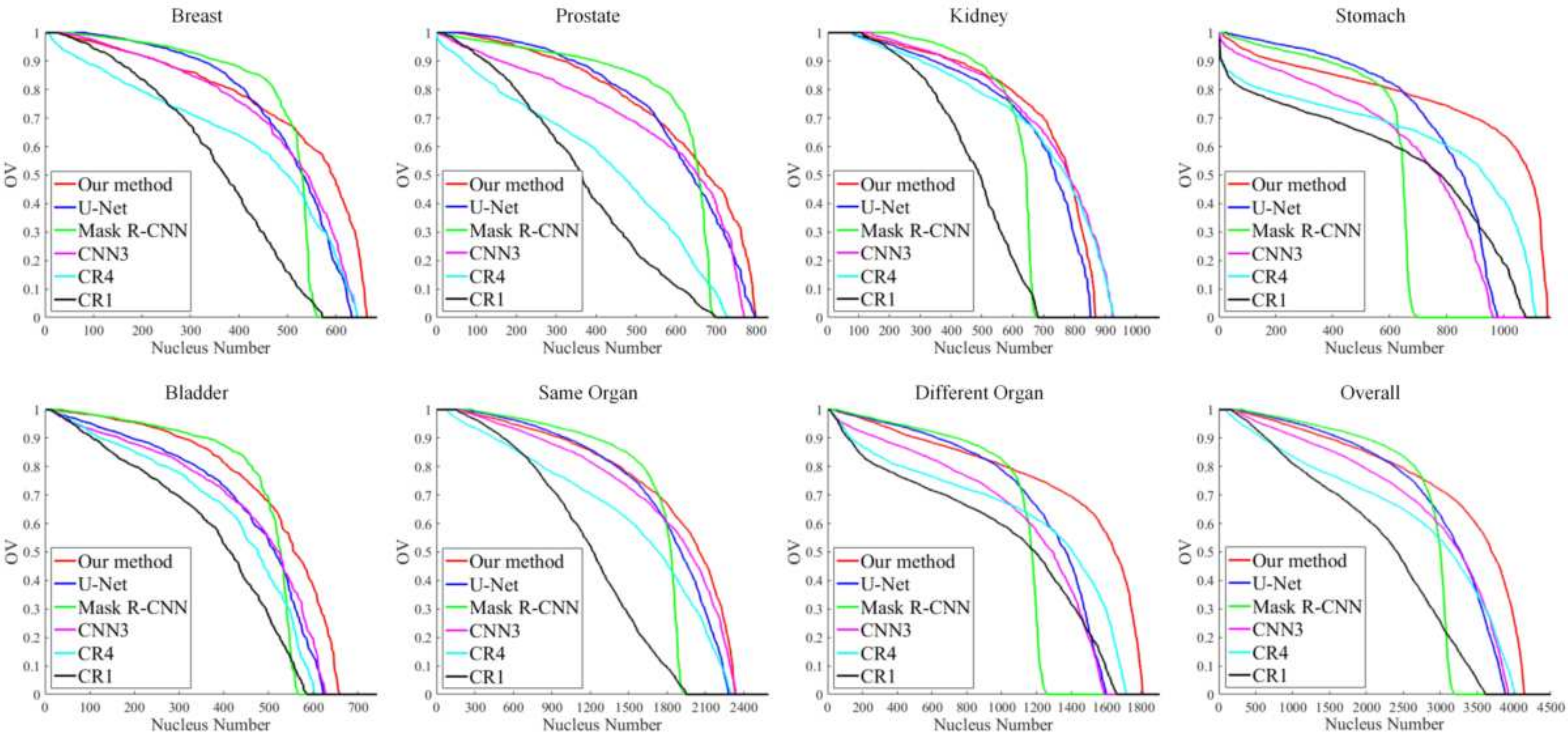}}
\caption{\footnotesize{Overlap evaluation. OV is the fraction of points on the ground truth nucleus marked as true positives, the larger the better.}}
\end{figure*}

\begin{figure*}[!t]
\renewcommand{\figurename}{\footnotesize{Fig.}}
\renewcommand{\captionlabeldelim}{.}
\centering
\noindent\makebox[\textwidth][c] {
\includegraphics[width=18cm]{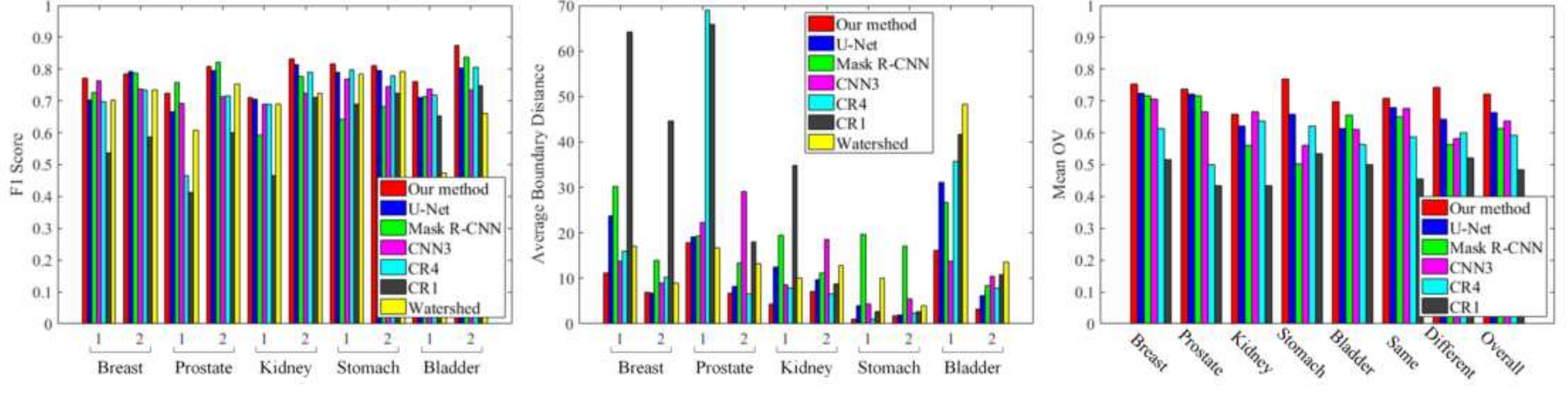}}
\caption{\footnotesize{Local and global comparative analyses for individual images from the testing sets and all methods. For the last column, the overlap values are sorted into decreasing order and then averaged over them.}}
\end{figure*}

\subsection{Local Pixel-Wise Evaluation}
With the training, model selection and testing procedure described in this section, we quantitatively compared all the state-of-the-arts. In Table IV and Figure 10, it can be clearly seen that the improvements obtained from using Sc${{\rm{D}}^2}$TE on three per pixel metrics are significant in comparison to other competing methods, especially for \emph{different organ} case. The multi-level feature maps learned by any of the Sc${{\rm{D}}^2}$TE layers (i.e. CSC modules) can be accessed by all subsequent layers. This encourages feature map reuse throughout the model, and leads to more accurate model. Another factor for the improved accuracy of Sc${{\rm{D}}^2}$TE is that the predictive performance can be boosted at each layer. By receiving additional supervision from all the preceding decision trees, the decision tree ensemble module monotonically decreases the errors. Specifically, the overall performance for the model without feature map reuse (e.g. CR1) is only 0.613. After the part feature map reuse, the performance can increase to 0.7195 (CR4), 0.7336 (Mask R-CNN), and 0.7583 (U-Net). Our fully dense feature map reuse with progressive predictions can further improve the performance to 0.7896. Also, deeply releasing the potentials of feature map reuse seems to make the boundary delineations more consistent as described by the reduced ABD value (pixels). In addition, the quantitative analysis on the dataset also reflects that, except for the watershed method, the encoder-decoder structures show a higher pixel-wise accuracy. It is interesting to compare their solutions to this segmentation problem with the other learning-based ones because they are computationally much more efficient.

\subsection{Global Segmentation Evaluation}
We randomly sampled and added a fixed number of nuclei from the ground truth, generated the segmented ones using different learning-based methods, and finally computed the corresponding OV values of them. Figure 9 shows the OV results for all the organs in the dataset. Our approach is more robust when used to segment the nuclei in digital pathology images and also competitive on the small values of the nucleus number. The accuracy of our approach remains higher for large values of the nucleus number, while the performance of the other methods decrease. In addition, our approach produces less zero values. All of these properties account for lower missing errors. From the last column of Figure 10, we see that our Sc${{\rm{D}}^2}$TE still dose best in terms of mean OV values but the others without or with partly reusing feature maps do worse, especially in \emph{different organ} case, due to the presence of heavy background clutter. This confirms the importance more contextual information to solve the problem.

\begin{figure}[!t]
\renewcommand{\figurename}{\footnotesize{Fig.}}
\renewcommand{\captionlabeldelim}{.}
\centering
\includegraphics[width=8.5cm]{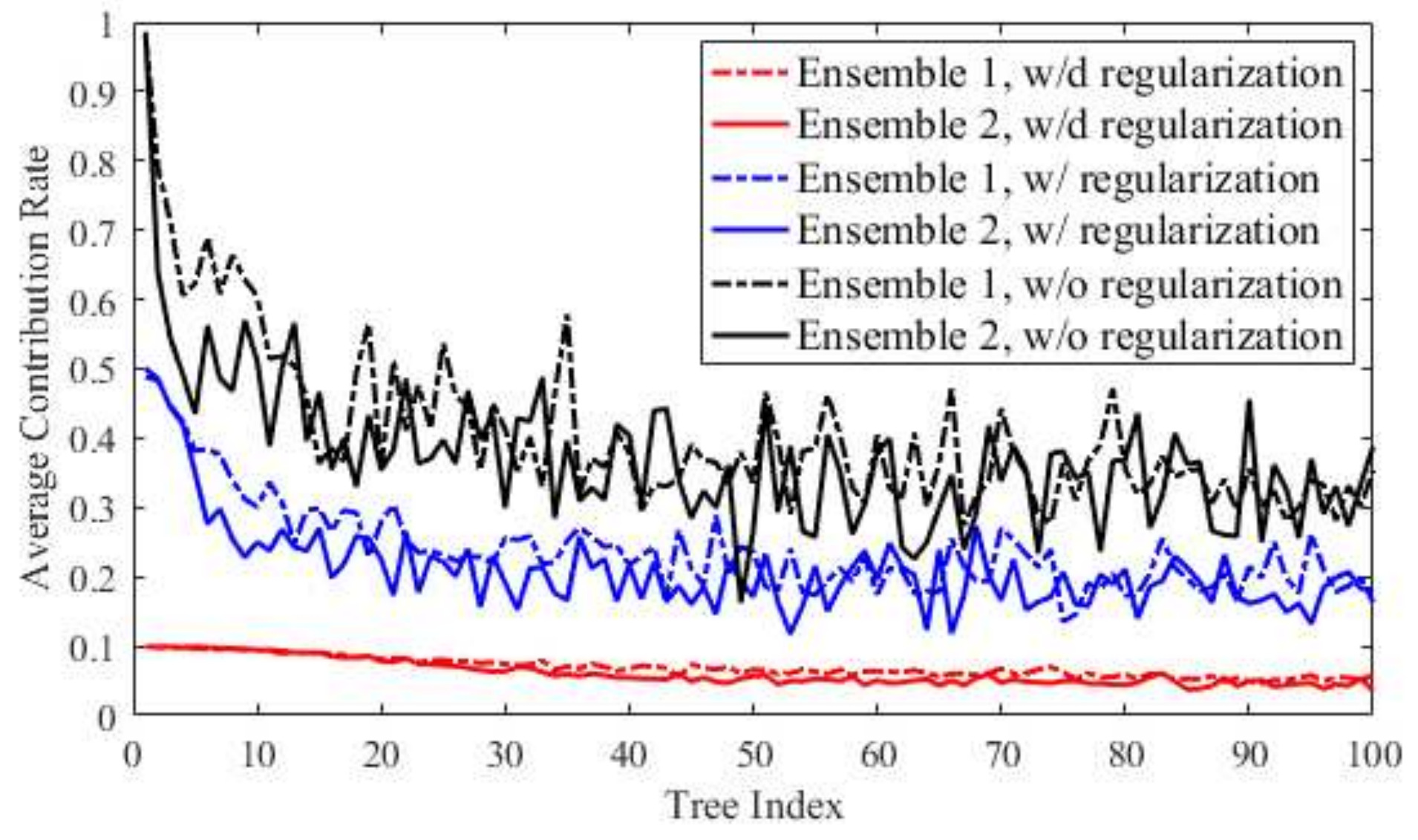} \\
\vspace{-0.25cm}
\caption{\footnotesize{The average contribution of decision trees in the decision tree ensemble module for different ${\ell _2}$ regularizations.}}
\end{figure}

\begin{table}[!t]
\renewcommand\arraystretch{1.2}
\renewcommand{\captionfont}{\footnotesize}
\renewcommand{\captionlabeldelim}{}
\begin{center}
\caption {}
{\textsc{A Comparison of Our Model with Different Feature Reuse}}
\vspace{0.2cm}
\setlength{\tabcolsep}{1pt}
\begin{tabular}{lcccccccc}
\hline
\multirow{2}*{Method} & \multicolumn{2}{c}{$1^{\rm{st}}$ layer} & \multicolumn{2}{c}{$2^{\rm{nd}}$ layer} & \multicolumn{2}{c}{$3^{\rm{rd}}$ layer} & \multicolumn{2}{c}{$4^{\rm{th}}$ layer}\\
\cmidrule(r){2-3} \cmidrule(r){4-5} \cmidrule(r){6-7} \cmidrule(r){8-9}
 & Time(s) & F1  & Time(s) & F1 & Time(s) & F1 & Time(s) & F1 \\
\hline
w/o feature reuse & 14.31 & 0.628 & 9.39 & 0.654 & 8.82 & 0.608 & 9.24 & 0.599 \\
w/ feature reuse & 14.61 & 0.632 & 25.51 & 0.661 & 25.68 & 0.701 & 25.66 & 0.732 \\
w/d feature reuse & 14.82 & 0.633 & 25.48 & 0.678 & 72.39 & 0.730 & 133.84 & 0.790 \\
\hline
\end{tabular}
\label{tab5}
\end{center}
\vspace{-0.5cm}
\end{table}

\subsection{Ablation Experiments on Sc${{D}^2}$TE}
It is also interesting to examine the behaviour of the proposed Sc${{\rm{D}}^2}$TE as the training and model setting are varied. It is worth noting that all the key hyper-parameters of comparing models are optimal and the time refers to the training time for each part. To explain the effectiveness of proposed method, we conducted the following comparisons:

\subsubsection{Impact analysis of the concatenated regularization}
As seen in Figure 11, the traditional tree ensemble learning model starts with a single or two decision trees that make significant contributions. This negatively affects the performance of the whole module on unseen data, makes the module over-sensitive to the contributions of these few, initially added trees, and might be prone to overfitting. By adopting a concatenated ${\ell _2}$ regularization approach to flatten the values of leaves, the expected contributions of the decision trees added in later iterations do not drop much, and thus the overfitting problem can be avoided. In addition, the penalty for maximum leaf size further improve the generalization performance of ensemble.

\subsubsection{Comparison analysis of feature map reuse methods}
To further justify the effectiveness of our Sc${{\rm{D}}^2}$TE (refers to w/d feature map reuse method) compared to w/o and w/ feature map reuse methods, we also trained two corresponding models to segment the nuclei. We used the same features and the same parameters used for the Sc${{\rm{D}}^2}$TE. The only differences are the connectivity patterns. Table V lists the results. As seen, the proposed method achieved the best balance between fitting quality and computational cost. We also observe that our fully dense connections have a regularizing effect, and with the increase of model depth, the greatly improved performance can be obtained. By contrast, the w/o feature map reuse method weakens the contributions of the decision trees learned in later iterations and leads to overfitting. As expected the performance gets worse as the depth increases, and the F1 score starts decreasing when the depth is larger than 2. Although the w/ feature map reuse method improves the segmentation result partially, it is still hard to obtain accurate pixel-wise prediction.

\begin{table}[!t]
\renewcommand\arraystretch{1.2}
\renewcommand{\captionfont}{\footnotesize}
\renewcommand{\captionlabeldelim}{}
\begin{center}
\caption {}
\textsc{A Comparison of Our Model with Different Types of Features as Inputs.}
\vspace{0.2cm}
\setlength{\tabcolsep}{2.5pt}
\begin{tabular}{lcccccc}
\hline
\multirow{2}*{Input features} & \multicolumn{2}{c}{$1^{\rm{st}}$ layer} & \multicolumn{2}{c}{$2^{\rm{nd}}$ layer} & \multicolumn{2}{c}{$3^{\rm{rd}}$ layer}\\
\cmidrule(r){2-3} \cmidrule(r){4-5} \cmidrule(r){6-7}
 & Time(s) & F1 & Time(s) & F1 & Time(s) & F1 \\
\hline
Fast CSC & 42.34 & 0.633 & 53.97 & 0.678 & 52.03 & 0.730 \\
Handcrafted & 39.54 & 0.477 & 53.21 & 0.566 & 50.81 & 0.536 \\
\hline
\end{tabular}
\label{tab7}
\end{center}
\vspace{-0.5cm}
\end{table}



\subsubsection{Multilayer Fast CSC features versus multilayer fast handcrafted features}
Finally, to test the representation ability of our Sc${{\rm{D}}^2}$TE, we compared our fast CSC approach with the handcrafted method, as summarized in Table VI. Here, we used the filters provided by Leung and Malik [\hyperlink{b48}{48}] to extract the handcrafted features as the input to each decision tree ensemble module. As seen, our approach is significantly more accurate than the handcrafted method. With the increase of model depth, higher layers of learning-based representations obtained by composing lower-level and high-level ones amplify aspects of the input that are important for discrimination and suppress irrelevant variations. A concatenation with multi-level context features to clarify the ambiguities further improved the model to capture the target details. In addition, fast CSC provides one way to reduce the effective complexity of the model by using the local-global decomposition, and thereby speeds up the execution time of the convolutions.

\section{Conclusion}
We proposed a sparse coding driven deep decision tree ensembles (Sc${{\rm{D}}^2}$TE) model for digital pathology image segmentation. In contrast to the state-of-the-arts, our approach has four advantages: 1) Sc${{\rm{D}}^2}$TE fully reuses the features of different levels through a sophisticated connectivity pattern, yielding condensed model that are easy to train, highly parameter-efficient, and without back-propagation; 2) Sc${{\rm{D}}^2}$TE introduces the fast CSC as the feature detector in each layer, which benefits from having a local shift invariant structure and can efficiently find a global sparsest representation of the whole image or maps; 3) A local encoder-decoder mechanism is proposed to boost the predictive performance at each layer, in which each decision tree seeks a solution that minimizes the regularized loss and improved performance can be obtained by combining all the preceding optimal decision trees through summation; 4) Experimental results on the digital pathology dataset demonstrate its superior performance to the state-of-the-arts. In the future, we would expect comparable performance, particularly for the case of more general application to the field of image processing.

\appendices

\ifCLASSOPTIONcaptionsoff
  \newpage
\fi




\begin{thebibliography}{1}
\bibitem{b1} \hypertarget{b1}
K.-H.~Yu, C.~Zhang, G.~J.~Berry, R.~B.~Altman \emph{et al.}, "Predicting non-small cell lung cancer prognosis by fully automated microscopic pathology image features," \emph{Nat. Commun.}, vol. 7, no. 12474, pp. 1--10, Aug. 2016.

\bibitem{b2} \hypertarget{b2}
Y.~Chen, Q.~Shen, S.~L.~White, Y.~Gokmen-Polar \emph{et al.}, "Three-dimensional imaging and quantitative analysis in CLARITY processed breast cancer tissues," \emph{Sci. Rep.}, vol. 9, no. 5624, pp. 1--13, Apr., 2019.

\bibitem{b3} \hypertarget{b3}
Z.~Tang, K.~V.~Chuang, C.~DeCarli, L.-W.~Jin \emph{et al.}, "Interpretable classification of Alzheimer's disease pathologies with a convolutional neural network pipeline," \emph{Nat. Commun.}, vol. 10, no. 2173, pp. 1--14, May 2019.

\bibitem{b4} \hypertarget{b4}
H.~Irshad, A.~Veillard, L.~Roux, and D.~Racoceanu, "Methods for nuclei detection, segmentation, and classification in digital histopathology: A review---Current status and future potential," \emph{IEEE Rev. Biomed. Eng.}, vol. 7, pp. 97--114, Apr. 2014.

\bibitem{b5} \hypertarget{b5}
F.~Xing and L.~Yang, "Robust nucleus/cell detection and segmentation in digital pathology and microscopy images: A comprehensive review," \emph{IEEE Rev. Biomed. Eng.}, vol. 9, pp. 234--263, Jan. 2016.

\bibitem{b6} \hypertarget{b6}
F.~Xing, Y.~Xie, H.~Su, F.~Liu \emph{et al.}, "Deep learning in microscopy image analysis: A survey," \emph{IEEE Trans. Neural Netw. Learn. Syst.}, vol. 29, no. 10, pp. 4550--4568, Oct. 2018.

\bibitem{b7} \hypertarget{b7}
J.~Song, L.~Xiao, and Z.~Lian, "Boundary-to-marker evidence-controlled segmentation and MDL-based contour inference for overlapping nuclei," \emph{IEEE J. Biomed. Health Inform.}, vol. 21, no. 2, pp. 451--464, Mar. 2017.


\bibitem{b8} \hypertarget{b8}
S.~Ali and A.~Madabhushi, "An integrated region-, boundary-, shape-based active contour for multiple object overlap resolution in histological imagery," \emph{IEEE Trans. Med. Imag.}, vol. 31, no. 7, pp. 1448--1460, Jul. 2012.

\bibitem{b9} \hypertarget{b9}
C.~Park, J.~Z.~Huang, J.~X.~Ji, and Y.~Ding, "Segmentation, inference and classification of partially overlapping nanoparticles," \emph{IEEE Trans. Pattern Anal. Mach. Intell.}, vol. 35, no. 3, pp. 669--681, Mar. 2013.

\bibitem{b10} \hypertarget{b10}
F.~Dong, H.~Irshad, E.-Y.~Oh, M.~F.~Lerwill \emph{et al.}, "Computational pathology to discriminate benign from malignant intraductal proliferations of the breast," \emph{PLoS ONE}, vol. 9, no. 12, p. e114885, Dec. 2014.

\bibitem{b11} \hypertarget{b11}
J.~Xu, L.~Xiang, Q.~Liu, H.~Gilmore \emph{et al.}, "Stacked sparse autoencoder (SSAE) for nuclei detection on breast cancer histopathology images," \emph{IEEE Trans. Med. Imag.}, vol. 35, no. 1, pp. 119--130, Jan. 2016.

\bibitem{b12} \hypertarget{b12}
F.~Xing, Y.~Xie, and L.~Yang, "An automatic learning-based framework for robust nucleus segmentation," \emph{IEEE Trans. Med. Imag.}, vol. 5, no. 2, pp. 550--566, Feb. 2016.

\bibitem{b13} \hypertarget{b13}
N.~Kumar, R.~Verma, S.~Sharma, S.~Bhargava \emph{et al.}, "A dataset and a technique for generalized nuclear segmentation for computational pathology," \emph{IEEE Trans. Med. Imag.}, vol. 36, no. 7, pp. 1550--1560, Jul. 2017.

\bibitem{b14} \hypertarget{b14}
L.~Zhang, M.~Sonka, L.~Lu, R.~M.~Summers \emph{et al.}, "Combining fully convolutional networks and graph-based approach for automated segmentation of cervical cell nuclei," in \emph{Proc. 14th IEEE Int. Symp. Biomed. Imag., Nano Macro}, Apr. 2017, pp. 406--409.

\bibitem{b15} \hypertarget{b15}
O.~Ronneberger, P.~Fischer, and T.~Brox, "U-net: Convolutional networks for biomedical image segmentation," in \emph{Proc. Int. Conf. Med. Image Comput. Comput.-Assist. Intervent.}, Nov. 2015, vol. 9351, pp. 234--241.

\bibitem{b16} \hypertarget{b16}
K.~He, G.~Gkioxari, P.~Doll$\acute{\rm {a}}$r, and R.~B.~Girshick, "Mask R-CNN," in \emph{Proc. IEEE Int. Conf. Comput. Vis.}, Oct. 2017, pp. 2980--2988.

\bibitem{b17} \hypertarget{b17}
P.~Naylor, M.~La$\acute{\rm {e}}$, F.~Reyal, and T.~Walter, "Segmentation of nuclei in histopathology images by deep regression of the distance map," \emph{IEEE Trans. Med. Imag.}, vol. 38, no. 2, pp. 448--459, Feb. 2019.

\bibitem{b18} \hypertarget{b18}
C.~Szegedy, W.~Liu, Y.~Jia, P.~Sermanet \emph{et al.}, "Going deeper with convolutions," in \emph{Proc. IEEE Conf. Comput. Vis. Pattern Recognit.}, Jun. 2015, pp. 1--9.

\bibitem{b19} \hypertarget{b19}
K.~He, X.~Zhang, S.~Ren, and J.~Sun, "Deep residual learning for image recognition," in \emph{Proc. IEEE Conf. Comput. Vis. Pattern Recognit.}, Jun. 2016, pp. 770--778.

\bibitem{b20} \hypertarget{b20}
G.~Huang, Z.~Liu, and L.~van~der~Maaten, "Densely connected convolutional networks," in \emph{Proc. IEEE Conf. Comput. Vis. Pattern Recognit.}, Jul. 2017, pp. 2261--2269.

\bibitem{b21} \hypertarget{b21}
J.~Long, E.~Shelhamer, and T.~Darrell, "Fully convolutional networks for semantic segmentation," in \emph{Proc. IEEE Conf. Comput. Vis. Pattern Recognit.}, Jun. 2015, pp. 3431--3440.

\bibitem{b22} \hypertarget{b22}
S.~Ren, K.~He, R.~Girshick, and J.~Sun, "Faster R-CNN: Towards real-time object detection with region proposal networks," \emph{IEEE Trans. Pattern Anal. Mach. Intell.}, vol. 39, no. 6, pp. 1137--1149, Jun. 2017.

\bibitem{b23} \hypertarget{b24}
T.-Y.~Lin, P.~Doll$\acute{\rm {a}}$r, R.~Girshick, K.~He \emph{et al.}, "Feature pyramid networks for object detection," in \emph{Proc. IEEE Conf. Comput. Vis. Pattern Recognit.}, Jul. 2017, pp. 936--944.

\bibitem{b24} \hypertarget{b24}
A.~Sironi, B.~Tekin, R.~Rigamonti, V.~Lepetit \emph{et al.}, "Learning separable filters," \emph{IEEE Trans. Pattern Anal. Mach. Intell.}, vol. 37, no. 1, pp. 94--106, Jan. 2015.

\bibitem{b25} \hypertarget{b25}
M. Seyedhosseini, M. Sajjadi, and T. Tasdizen, "Image segmentation with cascaded hierarchical models and logistic disjunctive normal networks," in \emph{Proc. IEEE Int. Conf. Comput. Vis.}, Dec. 2013, pp. 2168--2175.

\bibitem{b26} \hypertarget{b26}
J. Song, L. Xiao, M. Molaei, and Z. Lian, "Multi-layer boosting sparse convolutional model for generalized nuclear segmentation from histopathology images," \emph{Knowl. Based Syst.}, vol. 176, pp. 40--53, Jul. 2019.

\bibitem{b27} \hypertarget{b27}
Y.~Zhou, H.~Chang, K.~E.~Barner, and B.~Parvin, "Nuclei segmentation via sparsity constrained convolutional regression," in \emph{Proc. 12th IEEE Int. Symp. Biomed. Imag., Nano Macro}, Apr. 2015, pp. 1284--1287.

\bibitem{b28} \hypertarget{b28}
A.~Sironi, V.~Lepetit, and P.~Fua, "Multiscale centerline detection by learning a scale-space distance transform," in \emph{Proc. IEEE Conf. Comput. Vis. Pattern Recognit.}, 2014, pp. 2697--2704.

\bibitem{b29} \hypertarget{b29}
A.~Sironi, E.~T$\ddot{\rm {u}}$retken, V.~Lepetit, and P.~Fua, "Multiscale centerline detection," \emph{IEEE Trans. Pattern Anal. Mach. Intell.}, vol. 38, no. 7, pp. 1327--1341, Jul. 2016.

\bibitem{b30} \hypertarget{b30}
J.~Song, L.~Xiao, and Z.~Lian, "Contour-seed pairs learning-based framework for simultaneously detecting and segmenting various overlapping cells/nuclei in microscopy images," \emph{IEEE Trans. Image Process.}, vol. 27, no. 12, pp. 5759--5774, Dec. 2018.

\bibitem{b31} \hypertarget{b31}
I.~Arganda-Carreras, V.~Kaynig, C.~Rueden, K.~W.~Eliceiri, \emph{et al.}, "Trainable weka segmentation: a machine learning tool for microscopy pixel classification," \emph{Bioinformatics}, vol. 33, no. 15, pp. 2424--2426, Aug. 2017.

\bibitem{b32} \hypertarget{b32}
Z.-H.~Zhou and J.~Feng, "Deep forest: Towards an alternative to deep neural networks," in \emph{Proc. Int. Joint Conf. Artif. Intell.}, Aug. 2017, pp. 3553--3559.

\bibitem{b33} \hypertarget{b33}
F.~Xing, H.~Su, J.~Neltner, and L.~Yang, "Automatic Ki-67 counting using robust cell detection and online dictionary learning," \emph{IEEE Trans. Biomed. Eng.}, vol. 61, no. 3, pp. 859--870, Mar. 2014.

\bibitem{b34} \hypertarget{b34}
H.~Su, F.~Xing, X.~Kong, Y.~Xie, \emph{et al.}, "Robust cell detection and segmentation in histopathological images using sparse reconstruction and stacked denoising autoencoders," in \emph{Proc. Int. Conf. Med. Image Comput. Comput.-Assist. Intervent.}, Nov. 2015, vol. 9351, pp. 383--390.

\bibitem{b35} \hypertarget{b35}
M.~D.~Zeiler, D.~Krishnan, G.~W.~Taylor, and R.~Fergus, "Deconvolutional networks," in \emph{Proc. IEEE Conf. Comput. Vis. Pattern Recognit.}, Jun. 2010, pp. 2528--2535.

\bibitem{b36} \hypertarget{b36}
H.~Bristow, A.~Eriksson, and S.~Lucey, "Fast convolutional sparse coding," in \emph{Proc. IEEE Conf. Comput. Vis. Pattern Recognit.}, Jun. 2013, pp. 391--398.

\bibitem{b37} \hypertarget{b37}
B.~Wohlberg, "Efficient convolutional sparse coding," in \emph{Proc. IEEE Int. Conf. Acoust., Speech Signal Process.}, 2014, pp. 7173--7177.

\bibitem{b38} \hypertarget{b38}
F.~Heide, W.~Heidrich, and G.~Wetzstein, "Fast and flexible convolutional sparse coding," in \emph{Proc. IEEE Conf. Comput. Vis. Pattern Recognit.}, Oct. 2015, pp. 5135--5143.

\bibitem{b39} \hypertarget{b39}
V.~Papyan, Y.~Romano, and J.~Sulam, "Convolutional dictionary learning via local processing," in \emph{Proc. IEEE Int. Conf. Comput. Vis.}, Oct. 2017, pp. 5296--5304.

\bibitem{b40} \hypertarget{b40}
J.~Sulam, V.~Papyan, Y.~Romano, and M.~Elad, "Multilayer convolutional sparse modeling: Pursuit and dictionary learning," \emph{IEEE Trans. Signal Process.}, vol. 66, no. 15, pp. 4090--4104, Aug. 2018.

\bibitem{b41} \hypertarget{b41}
T.~Hastie, R.~Tibshirani, and J.~Friedman, \emph{The Elements of Statistical Learning}, New York, NY, USA: Springer, 2001.

\bibitem{b42} \hypertarget{b42}
J.~Friedman, T.~Hastie, and R.~Tibshirani, "Additive logistic regression: A statistical view of boosting," \emph{The Annals of Statistics}, vol. 28, no. 2, pp. 337--407, 2000.

\bibitem{b43} \hypertarget{b43}
E.~Zisselman, J.~Sulam, and M.~Elad, "A local block coordinate descent algorithm for the CSC model," in \emph{Proc. IEEE Conf. Comput. Vis. Pattern Recognit.}, Jun. 2019, pp. 8208--8217.

\bibitem{b44} \hypertarget{b44}
T.~Chen and C.~Guestrin, "XGBoost: A scalable tree boosting system," in \emph{Proc. ACM SIGKDD Int. Conf. Knowl. Disc. Data Mining}, Aug. 2016, pp. 785--794.

\bibitem{b45} \hypertarget{b45}
{The Cancer Genome Atlas (TCGA)}, accessed on May 14, 2016. [Online]. Available: http://cancergenome.nih.gov/

\bibitem{b46} \hypertarget{b46}
M. Abadi, A. Agarwal, P. Barham, E. Brevdo \emph{et al.}, "Tensorflow: Large-scale machine learning on heterogeneous distributed systems," arXiv preprint arXiv:1603.04467, Mar. 2016.

\bibitem{b47} \hypertarget{b47}
J. Mairal, F. Bach and J. Ponce, "Task-driven dictionary learning," \emph{IEEE Trans. Pattern Anal. Mach. Intell.}, vol. 34, no. 4, pp. 791--804, Apr. 2012.

\bibitem{b48} \hypertarget{b48}
T. Leung and J. Malik, "Representing and recognizing the visual appearance of materials using three-dimensional textons," \emph{Int. J. Comput. Vis.}, vol. 43, no. 1, pp. 29--44, Jun. 2001.
\end{thebibliography}
\end{document}